\documentclass{article}

\usepackage[utf8]{inputenc} 
\usepackage[T1]{fontenc}    
\usepackage{url}            
\usepackage{booktabs}       
\usepackage{amsfonts}       
\usepackage{nicefrac}       

\usepackage{microtype}
\usepackage{graphicx}
\usepackage{booktabs} 

\usepackage{bbm}
\usepackage{url}
\usepackage{color}
\usepackage{amsthm}
\usepackage{booktabs}
\usepackage{amsmath, amssymb}
\usepackage{mathtools}
\usepackage{amsfonts}
\usepackage{algorithm}
\usepackage[noend]{algorithmic}
\usepackage{amssymb, amsmath,charter, latexsym}
\usepackage{enumerate}

\usepackage{geometry}
\usepackage{algorithm}
\usepackage[noend]{algorithmic}

\theoremstyle{remark}

\usepackage[hidelinks]{hyperref}
\usepackage[switch]{lineno}

\usepackage[english]{babel}

\usepackage[numbers]{natbib}
\usepackage{array}
\usepackage{colortbl}
\usepackage{enumitem}
\usepackage{multirow}
\usepackage{multicol}
\usepackage{caption}
\usepackage{color}
\usepackage{lipsum}
\usepackage{textcomp}
\usepackage{gensymb}

\DeclareMathOperator{\bi}{\mathbf{i}}
\DeclareMathOperator{\bo}{\mathbf{o}}

\DeclareMathOperator{\bn}{\mathbf{n}}

\newenvironment{tight_itemize}{
\begin{itemize}[leftmargin=15pt,nosep]
  \setlength{\topsep}{0pt}
  \setlength{\itemsep}{0pt}
  \setlength{\parskip}{0pt}
  \setlength{\parsep}{0pt}
}{\end{itemize}}

\newenvironment{tight_enumerate}{
\begin{enumerate}[leftmargin=15pt,nosep]
  \setlength{\topsep}{0pt}
  \setlength{\itemsep}{0pt}
  \setlength{\parskip}{0pt}
  \setlength{\parsep}{0pt}
}{\end{enumerate}}

\newcommand{\para}[1]{\paragraph{#1}}

\newcommand{\webpage}[1]{\href{https://akshatdave.github.io/pandora}{#1}}
\newcolumntype{L}{>{\centering\arraybackslash}m{1.5cm}}
\newcolumntype{M}{>{\centering\arraybackslash}m{1.2cm}}

\newcommand\blfootnote[1]{%
  \begingroup
  \renewcommand\thefootnote{}\footnote{#1}%
  \addtocounter{footnote}{-1}%
  \endgroup
}
\providecommand{\keywords}[1]
{
  \small	
  \textbf{\textit{Keywords---}} #1
}

\title{PANDORA: \\ Polarization-Aided Neural Decomposition Of Radiance}
\author{Akshat Dave, Yongyi Zhao and Ashok Veeraraghavan  \\ ECE Department \\
Rice University, Houston, USA
\blfootnote{{\small Email: \href{mailto:akshat@rice.edu}{akshat@rice.edu}, Project Webpage: \webpage{akshatdave.github.io/pandora}}}}
\date{}

\begin{document}

\maketitle

\begin{abstract}%
Reconstructing an object's geometry and appearance from multiple images, also known as inverse rendering, is a fundamental problem in computer graphics and vision.  Inverse rendering is inherently ill-posed because the captured image is an intricate function of unknown lighting conditions, material properties and scene geometry. Recent progress in representing scene properties as coordinate-based neural networks have facilitated neural inverse rendering resulting in impressive geometry reconstruction and novel-view synthesis.  Our key insight is that polarization is a useful cue for neural inverse rendering as polarization strongly depends on surface normals and is distinct for diffuse and specular reflectance. With the advent of commodity, on-chip, polarization sensors, capturing polarization has become practical. Thus, we propose PANDORA, a polarimetric inverse rendering approach based on implicit neural representations. From multi-view polarization images of an object, PANDORA jointly extracts the object's 3D geometry, separates the outgoing radiance into diffuse and specular and estimates the illumination incident on the object. We show that PANDORA outperforms state-of-the-art radiance decomposition techniques. PANDORA outputs clean surface reconstructions free from texture artefacts, models strong specularities accurately and estimates illumination under practical unstructured scenarios. 
\\
\end{abstract}
\section{Introduction}
\begin{figure}
    \centering
    \includegraphics[width=\textwidth]{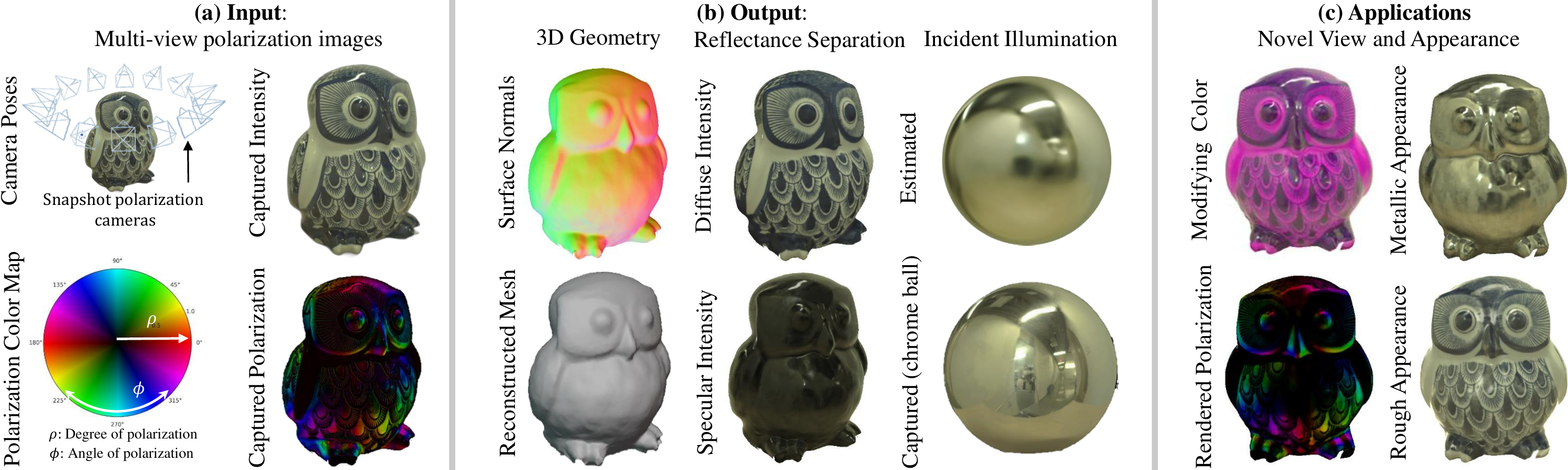}
    \caption{\textbf{PANDORA Overview:} PANDORA utilizes multi-view polarization images with known poses (a) and outputs the object's 3D geometry, separation of radiance in to diffuse and specular along with incident illumination (b). The learned PANDORA model can be applied to render the object under novel views and edit the object's appearance (c). Please refer to our \webpage{project webpage} for renderings of these outputs and applications under varying viewpoints.}
    \label{fig:teaser}
\end{figure}

Inverse rendering involves reconstructing an object's appearance and geometry from multiple images of the object captured under different viewpoints and/or lighting conditions. It is important for many computer graphics and vision applications such as re-lighting, synthesising novel views and blending real objects with virtual scenes. Inverse rendering is inherently challenging because the object's 3D shape, surface reflectance and incident illumination are intermixed in the captured images. A diverse array of techniques have been proposed to alleviate this challenge by incorporating prior knowledge about the scene, by optimizing the scene parameters iteratively using differentiable rendering and by using imaging modalities that exploit unique properties of light such as spectrum, polarization and time. 

\para{Neural Inverse Rendering.}
Recent works demonstrate that modelling the outgoing radiance and object shape as coordinate-based neural networks results in impressive novel-view synthesis (NeRF) \cite{mildenhall2020} and surface reconstruction (VolSDF) \cite{yariv2021volume} from multi-view captures. The outgoing radiance from the object is a combination of different components of surface reflectance and illumination incident on the object. As a result, separation and modification of components of the captured object's reflectance is not possible with works such as NeRF and VolSDF. Moreover, the diffuse and specular components of object reflectance have different view dependence. Using the same network to model a combination of difuse and specular radiance results in inaccurate novel view synthesis.

\para{Radiance Decomposition.}
Decomposition of the outgoing radiance into reflectance parameters and incident illumination is inherently ill-posed. Recent works such as PhySG \cite{physg2020} and NeuralPIL \cite{boss2021neural} aim to address the ill-posed nature of radiance decomposition by employing spherical Gaussians and data-driven embeddings respectively to model the reflectance and lighting. While these approach provides plausible decomposition in simple settings under large number of measurements, the decomposition is inaccurate in challenging scenarios such as strong specularities and limited views (Fig. \ref{fig:sim_radiance}) leading to blurrier specular reconstructions and artefacts in surface reconstruction. 

\para{Key idea: Polarization as a cue for reflectance decomposition.}
Our key insight is that polarimetric cues aid in radiance decomposition. Polarization has a strong dependence on surface normals. The diffuse and specular reflectance components have different polarimetric properties: the specular is more polarized than diffuse and the polarization angle for the two components are orthogonal. The advent of snapshot polarimetric cameras has made it pratical to capture this polarization information. In this work, we present our approach PANDORA that exploits multi-view polarization images for jointly recovering the 3D surface, separating the diffuse-specular radiance and estimating the incident illumination. 

\para{Our approach.}
PANDORA models the geometry as an implicit neural surface similar to VolSDF. Implicit coordinate based networks are used to model the reflectance properties. Incident lighting is modelled as an implicit network with integrated directional embeddings \cite{verbin2021ref}. We propose a differentiable rendering framework that takes as input the surface, reflectance parameters and illumination and renders polarization images under novel views. Given a set of multi-view polarization images, we jointly optimize parameters of the surface, reflectance parameters and incident illumination to minimize rendering loss. 

\para{Contributions.}
Our contributions are as follows:
\begin{tight_itemize}
\item Polarized neural rendering: We propose a framework to render polarization images from implicit representations of the object geometry, surface reflectance and illumination.

\item 3D surface reconstruction: Equipped with implicit surface representation and polarization cues, PANDORA outputs high quality surface normal, signed distance field and mesh representations of the scene.   

\item Diffuse-specular separation: We demonstrate accurate diffuse-specular radiance decomposition on real world scenes under unknown illumination. 

\item Incident illumination estimation: We show that PANDORA can estimate the illumination incident on the object with high fidelity.

\end{tight_itemize}

\para{Assumptions.}
In this work, we assume that the incident illumination is completely unpolarized. The object is assumed to be opaque and to be made up of dielectric materials such as plastics, ceramics etc as our polarimetric reflectance model doesn't handle metals. We focus on direct illumination light paths. Indirect illumination and self-occlusions are currently neglected. 

\section{Related Work}
\para{Inverse Rendering}. The goal of inverse rendering is to recover scene parameters from a set of associated images. Inverse rendering approaches traditionally rely on multi-view geometry \cite{schoenberger2016mvs,schoenberger2016sfm}, photometry \cite{barron2014shape} and structured lighting \cite{Nayar2006,OToole2016} for 3D reconstruction \cite{Vlasic2009}, reflectance separation \cite{Nayar2006,lin2002diffuse-specular}, material characterization \cite{Gkioulekas2013} and illumination estimation \cite{Ramamoorthi2001,Debevec1997}. Due to the ill-posed nature of inverse rendering, these approaches often require simplifying assumptions on the scene such as textured surfaces, Lambertian reflectance, direct illumination and simple geometry. Methods that aim to work in generalized scene settings involve incorporating scene priors \cite{kim2018inverse,yu19inverserendernet,Chen_2020_ECCV}, iterative scene optimization using differentiable rendering \cite{Li2018,Zhang2019DTRT} and exploiting different properties of light such as polarization \cite{Zhao2020}, time-of-flight \cite{yi2021differentiable} and spectrum \cite{li2021spectral}. 

\para{Neural Inverse Rendering}. Recent emergence of neural implicit representations \cite{xie2021neural} has led to an explosion of interest in neural inverse rendering \cite{tewari2021advances}. Neural implicit representations use a coordinate-based neural network to represent a visual signals such as images, videos, and 3D objects \cite{Sitzmann2019,Park2019,Mescheder2019}. These representations are powerful because the resolution of the underlying signal is limited only by the network capacity, rather than the discretization of the signal. Interest from the vision community originated largely due to neural radiance field (NeRF) \cite{mildenhall2020}, which showed that modelling radiance using implicit representations leads to high-quality novel view synthesis. 

Since the advent of NeRF, several works have exploited neural implicit representations for inverse rendering applications. IDR \cite{yariv2020multiview}, UNISURF \cite{oechsle2021unisurf}, NeuS \cite{wang2021neus} and VolSDF \cite{yariv2021volume} demonstrate state-of-the-art 3D surface reconstruction from multi-view images by extending NeRF's volume rendering framework to handle implicit surface representations. Accurate surface normals are crucial for modelling polarization and reflectance. Thus, we use ideas from one such work, VolSDF \cite{yariv2021volume}, as a build block in PANDORA.

 NeRF models the net outgoing radiance from a scene point in which both the material properties and the lighting are mixed. Several approaches such as NeRV \cite{nerv2021}, NeRD \cite{boss2021nerd}, NeuralPIL \cite{boss2021neural}, PhySG \cite{physg2020}, RefNeRF \cite{verbin2021ref} have looked at decomposing this radiance into reflectance components and illumination. PhySG and NeuralPIL employ spherical Gaussian and data-driven embeddings  to model the scene's illumination and reflectance. RefNeRF introduces integrated directional embeddings (IDEs) to model radiance from specular reflections and illumination and demonstrates improved novel view synthesis. Inspired from RefNeRF, we incorporate IDEs in our framework. Equipped with IDEs, implicit surface representation and polarimetric acquisition, PANDORA demonstrate better radiance decomposition than the state-of-the-art techniques, NeuralPIL and PhySG (Fig. \ref{fig:sim_radiance},\ref{fig:real_radiance}, Table \ref{table:metrics})  




\para{Polarimetric Inverse Rendering}. 
Polarization strongly depends on the surface geometry leading to several single view depth and surface normal imaging approaches \cite{Miyazaki2003,smith2018height,Ba2020,baek2018simultaneous,lei2021shape}. Inclusion of polarization cues has also led to enhancements in multi-view stereo \cite{Cui2017,Zhao2020,Fukao2021,ding2021polarimetric}, SLAM \cite{Yang2018} and time-of-flight imaging \cite{kadambi2015polarized}. The diffuse and specular components of reflectance have distinct polarization properties and this distinction has been utilized for reflectance seperation \cite{ma2007rapid,Cui2017,kim2016single}, reflection removal \cite{Lyu2019} and spatially varying BRDF estimation \cite{Deschaintre21}. PANDORA exploits these polarimetric cues for 3D reconstruction, diffuse-specular separation and illumination estimation. 

Traditionally acquiring polarization information required capturing multiple measurements by rotating a polarizer infront of the camera, unfortunately prohibiting fast acquisition. The advent of single-shot polarization sensors, such as the Sony
IMX250MZR (monochrome) and IMX250MYR (color) \cite{SonyPol} in commercial-grade off-the-shelf machine vision cameras has 
made polarimetric acquisition faster and more practical. These sensors have a grid of polarizers oriented at different angles attached
on the CMOS sensor enabling the capture of polarimetric cues at
the expense of spatial resolution. Various techniques have been
proposed for polarization and color demosaicking of the raw sensor
measurements \cite{morimatsu2020monochrome,qiu2021linear}. In PANDORA we use such a camera  FLIR BFS-U3-51S5P-C to capture polarization information for every view in a single shot. 



\section{Polarization as cue for radiance separation}
Here we introduce our key insight on how polarimetric cues aide in decomposing radiance into the diffuse and specular components. First we derive the polarimetric properties of diffuse and specular reflectance. Then we demonstrate how these cues can aide in separating the combined reflectance by the means of a simple scene.  
\begin{figure}[ht!]
    \centering
            \includegraphics[width=0.9\linewidth]{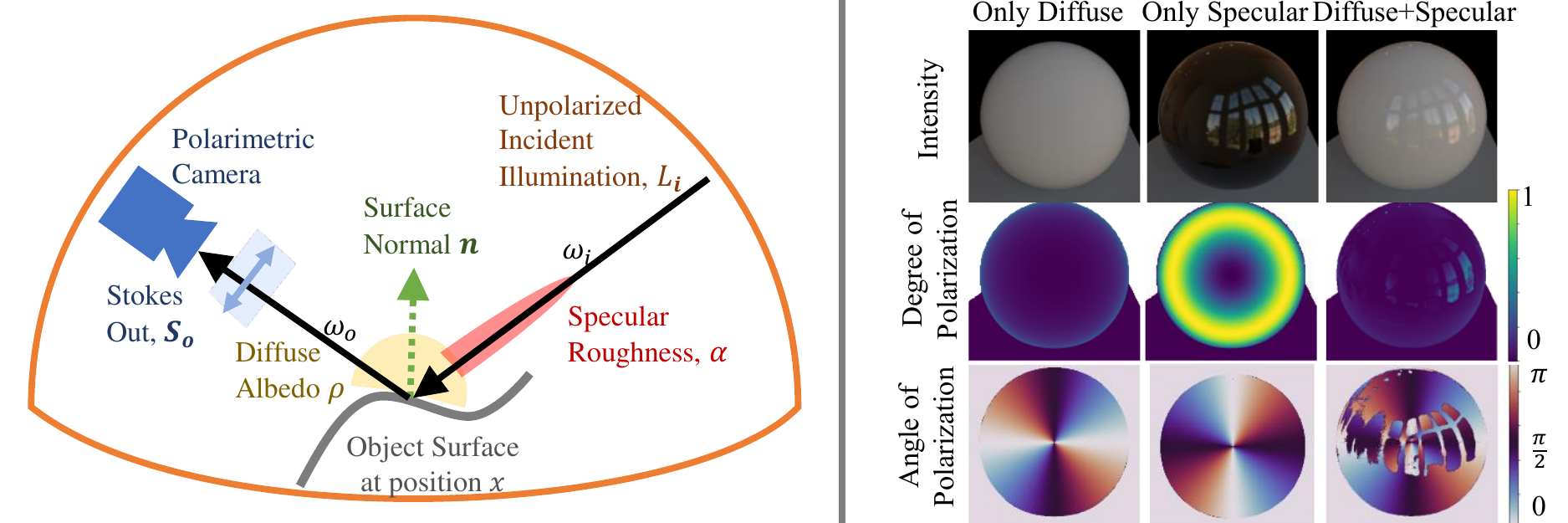}
    \caption{\textbf{Polarization as a cue for radiance decomposition} Left: Illustration of notations for our polarized image formation model. Right: Polarimetric cues for different radiance components. Diffuse radiance has a lower degree of polarization than the specular radiance. The polarization angles of diffuse and specular components are orthogonal. These polarization cues are used for better radiance decomposition.}
    \label{fig:pol_as_a_cue}
\end{figure}
\subsection{Theory of polarized reflectance}
\para{Stokes Vector.}
The polarization state of light ray at $\mathbf{x}$ along direction $\omega$ is modelled as Stokes vector comprising of four components, $S(\mathbf{x},\omega) = [S_0 \quad S_1 \quad S_2 \quad S_3]$ \cite{collett1992polarized}. We assume there is no circular polarization and thus neglect $S_3$. The Stokes vector can be parametrized as a function of three polarimetric cues: the total intensity, $L_o=S_0$, degree of polarization, $\beta_o=\sqrt{S_1^2+S_2^2}/S_0$ and angle of polarization, $\phi_o=\tan^{-1}(S_2/S_1)/2$. 

\para{Mueller Matrix.}
Upon interaction with an object's surface, the polarization state of light changes. The Stokes vector after the interaction can be modelled as the matrix multiplication of the input Stokes vector with a $4\times4$ matrix, known as the Mueller matrix.

\para{Polarimetric BRDF (pBRDF) model.}
The interaction of object for diffuse and specular components of the reflectance are different. The diffuse reflectance involves sub-surface scattering into the surface and then transmission out of the surface. The specular component can be modelled as a direct reflection from specular microfacets on the surface. 
The pBRDF model \cite{baek2018simultaneous} model these interactions as Mueller matrices for the diffuse and specular polarized reflectance, which we denoted as $H_d$ and $H_s$ respectively.  

\para{Incident Stokes vector}
Considering illumination is from far away sources, the dependance of $S_i$ on $\mathbf{x}$ can be dropped. Assuming the polarization to be completely unpolarized: 
\begin{align}
    S_i(\mathbf{x},\omega_i) = L_i(\mathbf{x},\omega_i)[1\quad0\quad0]^T \;,
\end{align}

\para{Exitant Stokes vector}
From the pBRDF model, the output Stokes vector at every point can be decomposed into the matrix multiplication of diffuse and specular Mueller matrices,$H_d$ and $H_s$, with the illumination Stokes vector $S_i$, 
\begin{align}
   S_o(\mathbf{x},\omega_i) = \int_{\Omega}{H_d \cdot S_i(\mathbf{x},\omega_i)d\omega} + \int_{\Omega}{H_s \cdot S_i(\mathbf{x},\omega_i)d\omega}
\end{align}
From $S_i$ and the pBRDF model, we derive that the outgoing Stokes vector at every point depends on the diffuse radiance $L_d$, specular reflectance $f_s$ and the incident illumination $L_i$ as,
\begin{align}
   S_o(\mathbf{x},\omega_i) = L_d    \begin{bmatrix}
        1 \\
        \beta_d(\theta_n) \cos(2\phi_n) \\
        -\beta_d(\theta_n) \sin(2\phi_n)\\
\end{bmatrix} + L_s
        \begin{bmatrix} 
         1\\
         \beta_s(\theta_n) \cos(2\phi_n) \\
         -\beta_s(\theta_n) \sin(2\phi_n)
       \end{bmatrix} \;, 
    \label{eq:S_o_L}
\end{align}
where we terms $\beta_d$/$\beta_s$ depend on Fresnel transmission/reflection coefficients for the polarization components parallel and perpendicular to the plane of incidence, $T^\parallel$/$R^\parallel$ and $T^\bot$/$R^\bot$
\begin{align}
    \beta_d = \frac{T^{\bot}-T^{\parallel}}{T^{\bot}+T^{\parallel}} \;, \quad
    \beta_s = \frac{R^{\bot}-R^{\parallel}}{R^{\bot}+R^{\parallel}} \;,
\end{align}
The Fresnel coefficients, $T$/$R$ solely depend on the elevation angle of the viewing ray with respect to the surface normal, $\theta_n = \cos^{-1}(\mathbf{n}\cdot\omega_o)$.
$\phi_o$ denotes the azimuth angle of the viewing ray with respect to the surface normals, $\phi_n = \cos^{-1}(\mathbf{n}_o,\mathbf{y}_o)$, where $\mathbf{n}_o$ is the normal vector perpendicular to the viewing ray and $\mathbf{y}_o$ is the $y$ axis of camera coordinate system.
    Please refer to Appendix \ref{sec:suppl_derivation} for the detailed derivation and functional forms for Fresnel coefficients.

Next we show how these polarimetric cues depend on the diffuse and specular reflectance and aid in radiance decomposition.
\subsection{Polarimetric properties of diffuse and specular radiance}
\label{sec:diff_spec_pol}
From Eq. \ref{eq:S_o_L}, the polarimetric cues of the captured Stokes vector are \begin{align*}
    L_o = L_d + L_s \;, \quad \beta_o = L_d \beta_d + L_s \beta_s \;, \quad \phi_o = \tan^{-1}(-\tan(2\phi_n))/2
\end{align*}
Fig. \ref{fig:pol_as_a_cue} shows polarimetric cues for a sphere scene for different reflectance properties. For only diffuse case (left), the degree of polarization increases with elevation angle and the angle of polarization is equal to the azimuth angle. For only specular case (middle), the degree of polarization increases with elevation angle until the Brewster's angle after which it reduces. The angle of polarization is shifted from azimuth angle by $90^\circ$. When both diffuse and specular reflectance are present (right), the polarimetric cues indicate if a region is dominated by diffuse or specular radiance. The specular areas have higher degree of polarization than diffuse areas. The two components have orthogonal angle of polarization.




\section{Our Approach}
\begin{figure*}[t!]
    \centering
    \includegraphics[width=1.0\linewidth]{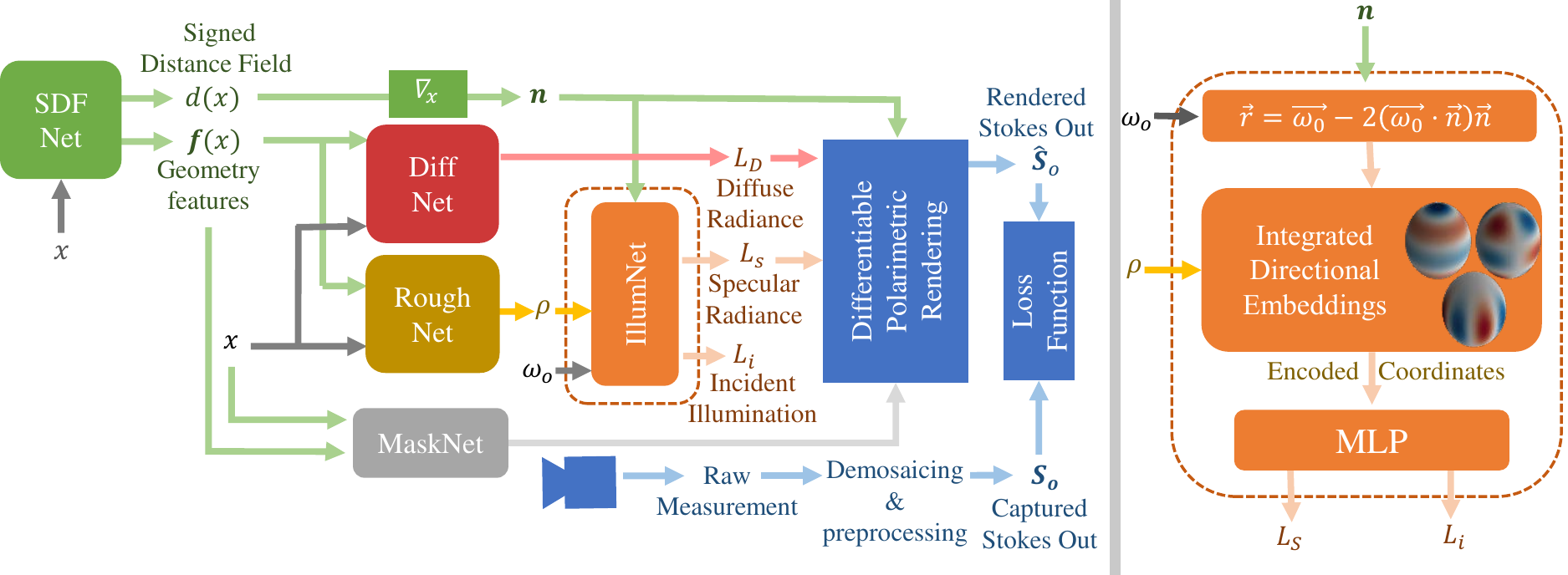}
    \caption{\textbf{PANDORA Pipeline}: Left: Our pipeline. We use coordinate-based networks to estimate surface normals, diffuse and specular radiance and incident illumination. From these parameters, we render exitant Stokes vector that is compared with captured Stokes vector and the loss is backpropagated to train the networks.Right: Detailed schematic of the Illumination Net}
    \label{fig:pipeline}
\end{figure*}
We aim to recover the object shape, diffuse and specular radiance along with the incident illumination from multi-view images captured from a consumer-grade snapshot polarization camera. Fig. \ref{fig:pipeline} summarizes our pipeline.
\subsection{Input}
\label{sec:approach_input}
PANDORA relies on the following inputs to perform radiance decomposition: 1) \textit{Polarization Images.} We capture multiple views around the object with a 4 MP snapshot polarization camera \cite{SonyPol} (Fig. \ref{fig:pipeline}(a)). These cameras comprise of polarization and Bayer filter arrays on the sensor to simultaneously capture color images for four different polarizer orientations at the expense of spatial resolution. We employ the demosaicking and post-processing techniques utilized in \cite{Zhao2020} to convert the raw sensor measurements into 4 MP RGB Stokes vector images. 2) \textit{Camera poses.} We use COLMAP Structure-from-motion technique \cite{schoenberger2016mvs},\cite{schoenberger2016sfm} to calibrate the camera pose from the intensity measurements of the polarization images. Thus for any pixel in the captured images, the camera position $\mathbf{o}$ and camera ray direction $\mathbf{d}$ are known.
An optional binary mask can also be used to remove signal contamination from the background. To create masks for real-world data, We use an existing object segmentation approach \cite{qin2020u2} for creating the object masks. The binary mask values are denoted as, $M(\mathbf{o},\mathbf{d})$.
\subsection{Implicit Surface estimation}
The Stokes vector measured by camera ray given by $\mathbf{o}$ and $\mathbf{d}$, the ray is sampled at $\mathcal{T}$ points. For a sample on the ray with travel distance $t$, its location is denoted at $\mathbf{r}(t) = \mathbf{o}+t\mathbf{d}$. The Stokes vector contribution of this sample depends on the scene opacity, $\sigma(\mathbf{r}(t))$ and exitant Stokes vector $S_o(\mathbf{r}(t),\mathbf{d})$. The observed Stokes vector, $S(\mathbf{o},\mathbf{d})$ is denoted by the integral, 
\begin{align}
    S(\mathbf{o},\mathbf{d}) = \int_{0}^{\infty}{T(t)\sigma(\mathbf{r}(t))S_o(\mathbf{r}(t),\mathbf{d})} dt \;,
    \label{eq:vol_render_int}
\end{align}
where $T(t) = exp\left(-\int_{0}^{\infty}{\sigma(\mathbf{r}(t))}\right)$ is the probablity that the ray travels to $t$ without getting occluded.

For rendering surfaces, the ideal opacity should have a sharp discontinuity at the ray surface intersection. Thus accurately sampling $\mathcal{T}$ points and consequently reconstructing sharp surfaces is challenging. High quality surface estimation is crucial for our approach as the polarization cues depend on the surface normals, (Eq. \ref{eq:S_o_L}). 

VolSDF \cite{yariv2021volume} has demonstrated significant improvements in surface estimation by modelling the signed distance field $d$ with a coordinate-based neural network. The opacity is then estimated as $\sigma(x) = \alpha \psi_{\beta}(-d(x))$ where $\alpha$,$\beta$ are learnable parameters and $\Psi$ is the Cumulative Distribution Function of the Laplace distribution with zero mean and scale $\beta$. They also propose a better sampling algorithm for $\mathcal{T}$ points utilizing the SDF representation. We follow the same algorithm as VolSDF for opacity generation. Similar to VolSDF, our pipeline comprises of an MLP, which we term SDFNet, that takes as input the position $\mathbf{x}$ and outputs the signed distance field at that position $\mathbf{x}$ along with geometry feature vectors $\mathbf{f}$ useful for radiance estimation. The SDF model also provides us with surface normals, which are used in estimating specular radiance and polarimetric cues.
\subsection{Neural Rendering Architecture}
\para{Diffuse Radiance Estimation.} Diffuse radiance is invariant of the viewing direction and only depends on the spatial location. The geometry features from SDFNet and the position are passed through another coordinate-based MLP, denoted as DiffuseNet, to output the diffuse radiance at that position $L_D(\mathbf{x})$.

\para{Specular Radiance Estimation.} Unlike the diffuse radiance, the specular radiance depends on the viewing angle $\mathbf{d}$ and the object roughness $\alpha{\mathbf{x}}$. First we estimate the object roughness using an coordinate-based MLP, RoughNet, similar in architecture to the DiffuseNet. 
For a certain object roughness, the obtained specular radiance involves integrating the specular BRDF along an incident direction factored by the incident illumination \cite{barron2021mipnerf}, which is a computationally expensive procedure that generally requires Monte Carlo. Inspired by \cite{verbin2021ref}, we instead use an IDE-based neural network to output the specular radiance, $L_D$ from the estimated roughness, $\alpha$ and surface normals, $\mathbf{n}$. Moreover,on setting roughness close to zero, IllumNet also provides us the incident illumination, $L_i$. 

\para{Volumetric Masking}
We exploit object masks to ensure only the regions in the scene corresponding to the target object are used for radiance decomposition. Even when the background is zero, VolSDF estimates surface normals which have to be masked out to avoid incorrect quering of the IllumNet. Rather than using the 2D masks on the rendered images, we found that learning a 3D mask of the target object helps in training, especially in the initial interations. This 3D mask $m(\mathbf{x})$ is 1 only for the positions $\mathbf{x}$ that the object occupies and represent's the object's visual hull. We use this 3D mask to obtain the diffuse and specular radiance that is clipped to zero at background values,
\begin{align}
L^m_D(\mathbf{x}) = m(\mathbf{x})\cdot L_D(\mathbf{x}) \quad
L^m_S(\mathbf{x},\mathbf{d}) = m(\mathbf{x})\cdot L_S(\mathbf{x},\mathbf{d}) 
\end{align}
The 3D mask is estimated using a coordinate-based MLP that we term MaskNet. This network is trained with the supervision of the input 2D object masks under different views. Similar to Eq. \ref{eq:vol_render_int}, the 3D mask values are accumulated along the ray and compared to the provided mask $M$ using the binary cross entropy loss:
\begin{equation}
    \mathcal{L}_{\text{mask}} = \mathbb{E}_{\mathbf{o},\mathbf{l}}{ \text{BCE}\left(M(\mathbf{o},\mathbf{d}),\hat{M}(\mathbf{o},\mathbf{d})\right)} \;,
    \label{eq:bce_masknet_loss}
\end{equation}
where $\hat{M}(\mathbf{o},\mathbf{d}) = \int_{0}^{\infty}{T(t)\sigma(\mathbf{r}(t))m(\mathbf{r}(t))} dt \;$.

\para{Neural Polarimetric Rendering}
Using the masked diffuse $L^m_D$, masked specular $L^m_S$ and the estimated surface normals $\mathbf{n}$, we can render the outgoing Stokes vector, $S_o(\mathbf{x},\mathbf{d})$ from Eq. \ref{eq:S_o_L}. On integrating outgoing Stokes vectors for points along the ray according to Eq. \ref{eq:vol_render_int}, we obtain the rendered Stokes vector $\hat{S}(\mathbf{x},\mathbf{d})$.

\begin{figure*}[t!]
\centering
\includegraphics[width=\linewidth]{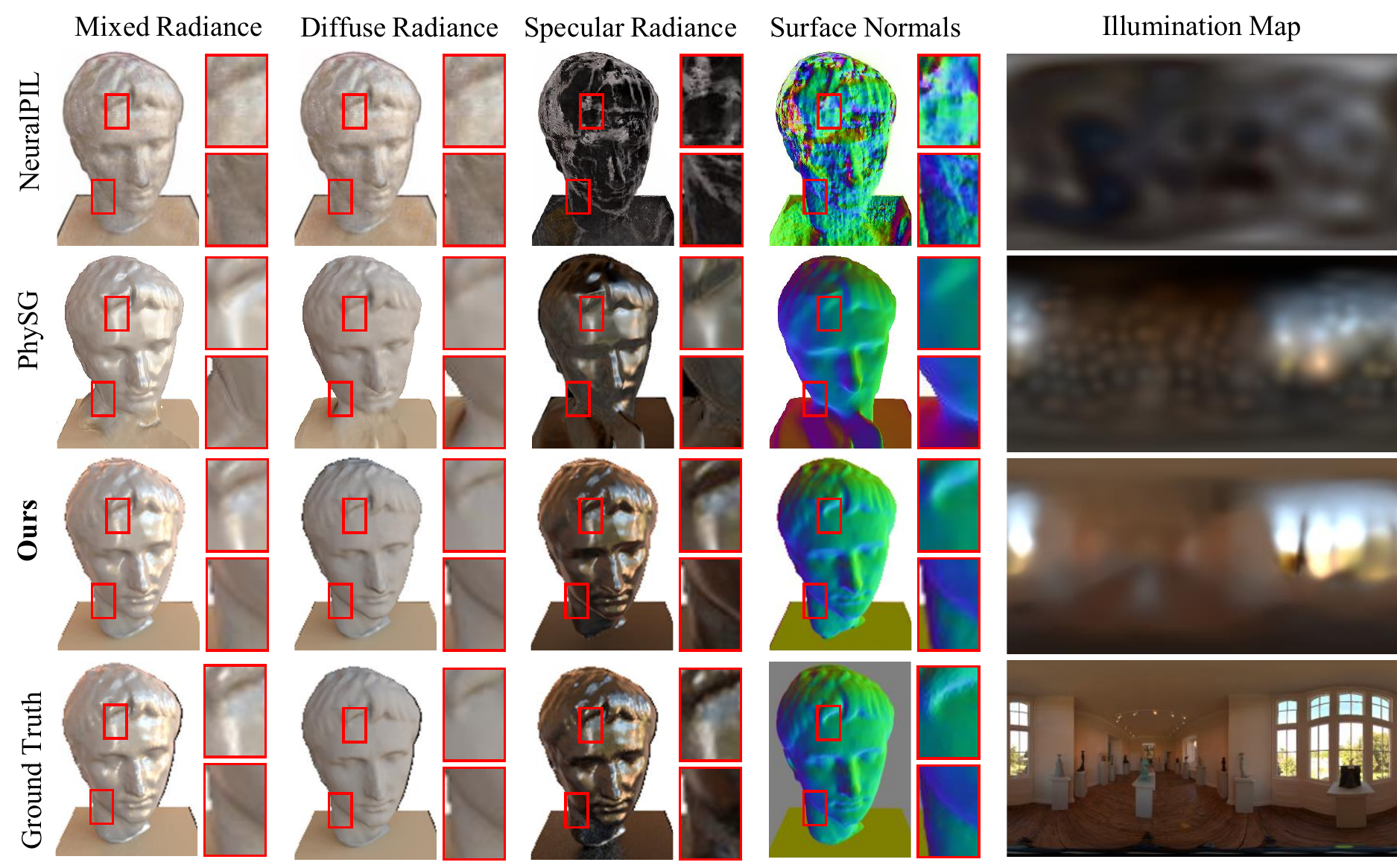}
\caption{\textbf{Comparison of reflectance separation and surface normals with baselines on rendered dataset}:NeuralPIL fails to estimate correct normals and illumination on this challenging scene with strong specularities and 45 views. PhySG exhibits blurrier speculars and illumination along with artifacts in the reconstructed normals. PANDORA outputs sharp specularities, cleaner surface and more accurate illumination.}
\label{fig:sim_radiance}
\end{figure*}
\subsection{Loss Function}
We compare the rendered Stokes vector $\hat{S} = [\hat{s_0},\hat{s_1},\hat{s_2}]^T$ with the captured Stokes vector $S = [s_0,s_1,s_2]^T$ (\S\ref{sec:approach_input}) using an L1 loss.The losss is masked to remove the effect of background values. The $s_1$ and $s_2$ could have low values in regions having low degree of polarization (Fig. \ref{fig:pol_as_a_cue}). We apply a weightage factor $w_s$ on the loss for $s_1$ and $s_2$ outputs to further encourage the network to consider polarimetric cues in the training. The Stokes loss is modelled as:
\begin{align}
   \mathcal{L}_{\text{stokes}} = \mathbb{E}_{\mathbf{o},\mathbf{l}}\left[M\cdot\| \hat{s_0} - {s_0}\|+w_s\cdot M \cdot\left(\|\hat{s_1} - s_1\|+\|\hat{s_2}-s_2\|\right)\right]
\label{eq:L_stokes}
\end{align}
    Additionally, similar to VolSDF \cite{yariv2021volume}, we have the Eikonal loss, $\mathcal{L}_{\text{SDF}}$ to encourage the SDFNet to approximate a signed distance field. 
\begin{align}
    \mathcal{L}_{\text{SDF}} = \mathbb{E}_{\mathbf{o},\mathbf{l}} \left(\|\mathbf{n}\|-1\right)^2
\end{align}
The net loss used to train all the networks described in the pipeline:
\begin{align}
    \mathcal{L}_{\text{net}} =  \mathcal{L}_{\text{stokes}} +0.1\mathcal{L}_{\text{SDF}} +\mathcal{L}_{\text{mask}}
\end{align}
\subsection{Implementation Details}
All the networks are standard MLPs with 4 layers each. SDFNet has 256 hidden units per layer and the other networks have 512 hidden units. ReLU activations are used in intermediate layers. Final activation in DiffNet and MaskNet and the final activation in IllumNet and RoughNet is softplus. 
Please refer to Appendix \ref{sec:suppl_details} for additional implementation details of our framework.


\section{Results and Evaluation }

\subsection{Datasets}
We generate the following datasets for evaluating radiance decomposition.
\begin{tight_enumerate}
    \item  Rendered Polarimetric Dataset (Fig. \ref{fig:sim_radiance}): Using Mitsuba2, we apply pBRDF on objects with complicated geometry and perform polarimetric rendering of multiple camera views under realistic environment lighting. 
    \item Real Polarimetric Dataset (Fig. \ref{fig:real_radiance},\ref{fig:real_diff_spec},\ref{fig:real_illum}): Using a snapshot polarimetric camera, we acquire multi-view polarized images of complex objects composed of materials with varying roughness, such as ceramics, glass, resin and plastic, under unstructured lighting conditions such as an office hallway. We also acquire the ground truth lighting using a chrome ball.
\end{tight_enumerate}
Please refer to Appendix \ref{sec:suppl_details} for additional details on the generation of these datasets and more examples from the datasets.
\begin{center}
\centering
\small
\begin{tabular}{|p{0.5 cm} c |m{1.6cm}|cc|cc|cc|c|c|}
 \hline
    \multicolumn{2}{|c|}{\multirow{3}{*}{Scene}}& {\multirow{3}{*}{Approach}} & \multicolumn{2}{c|}{Diffuse}& \multicolumn{2}{c|}{Specular}& \multicolumn{2}{c|}{Mixed} & Normals & Mesh\\
&&&  PSNR &  SSIM & PSNR & SSIM & PSNR &  SSIM & MAE & HD\\
&&&  $\uparrow$ (dB) &  $\uparrow$ &  $\uparrow$ (dB) & $\uparrow$ & $\uparrow$ (dB) &  $\uparrow$  & $\downarrow$ (\degree) & $\downarrow$ \\
 \hline
{\multirow{3}{*}{\rotatebox[origin=c]{90}{\hspace{0cm}Bust}}} &  \multirow{3}{*}{\includegraphics[width=1.cm,trim={0 0 0 0.2cm},clip]{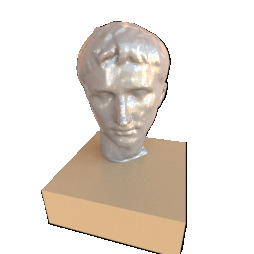}} & NeuralPIL&23.90&0.87&18.04&0.87&26.71&0.87&15.36&N/A\\
&&PhySG &22.64 & 0.94 & 23.00 & 0.94 & 19.94 & 0.72 & 9.81&0.012\\
&&Ours & 25.82 & 0.81 & 22.96 & 0.75 & 22.79 & 0.79 & 3.91&0.003\\
 \hline
{\multirow{3}{*}{\rotatebox[origin=c]{90}{\hspace{0cm}Sphere}}}
&   \multirow{3}{*}{\includegraphics[width=1.cm,trim={0 0 0 0.2cm}]{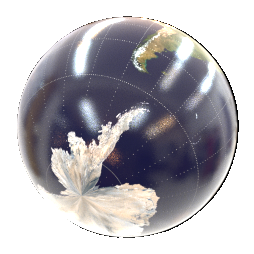}}
&NeuralPIL &13.09&0.55&12.92&0.55&20.04&0.66&38.73&N/A\\
&&PhySG& 21.76 & 0.76 & 18.90 & 0.76 & 17.93& 0.70& 8.42&0.011\\
&&Ours & 24.33 & 0.77 & 22.70 & 0.89 & 21.76 & 0.81 & 1.41&0.003\\
 \hline
    \end{tabular}
\captionof{table}{\textbf{Quantiative evaluation on rendered scenes} We evaluate PANDORA and state-of-the-art methods on held-out testsets of 45 images for two rendered scenes. We report the peak average signal-to-noise ratio (PSNR) and structured similarity (SSIM) of diffuse, specular and net radiance, mean angular error (MAE) of surface normals and the Hausdorff distance (HD) of the reconstructed mesh.PANDORA consistently outperforms state-of-the-art in radiance separation and geometry estimation.} 
\label{table:metrics}
\end{center}


\subsection{Comparisons with Baselines}
We demonstrate that PANDORA excels in 3D reconstruction, diffuse-specular separation and illumination estimation compared to two existing state-of-the-art radiance decomposition baselines, NeuralPIL\cite{boss2021neural} and PhySG \cite{physg2020}. These baselines cannot exploit polarization cues and are run on radiance-only images using the public code implementations provided by the authors. We then show additional applications of PANDORA and an ablation study to analyse the crucial aspects of our algorithm.

\begin{figure}[t!]
\centering
\includegraphics[width=\linewidth]{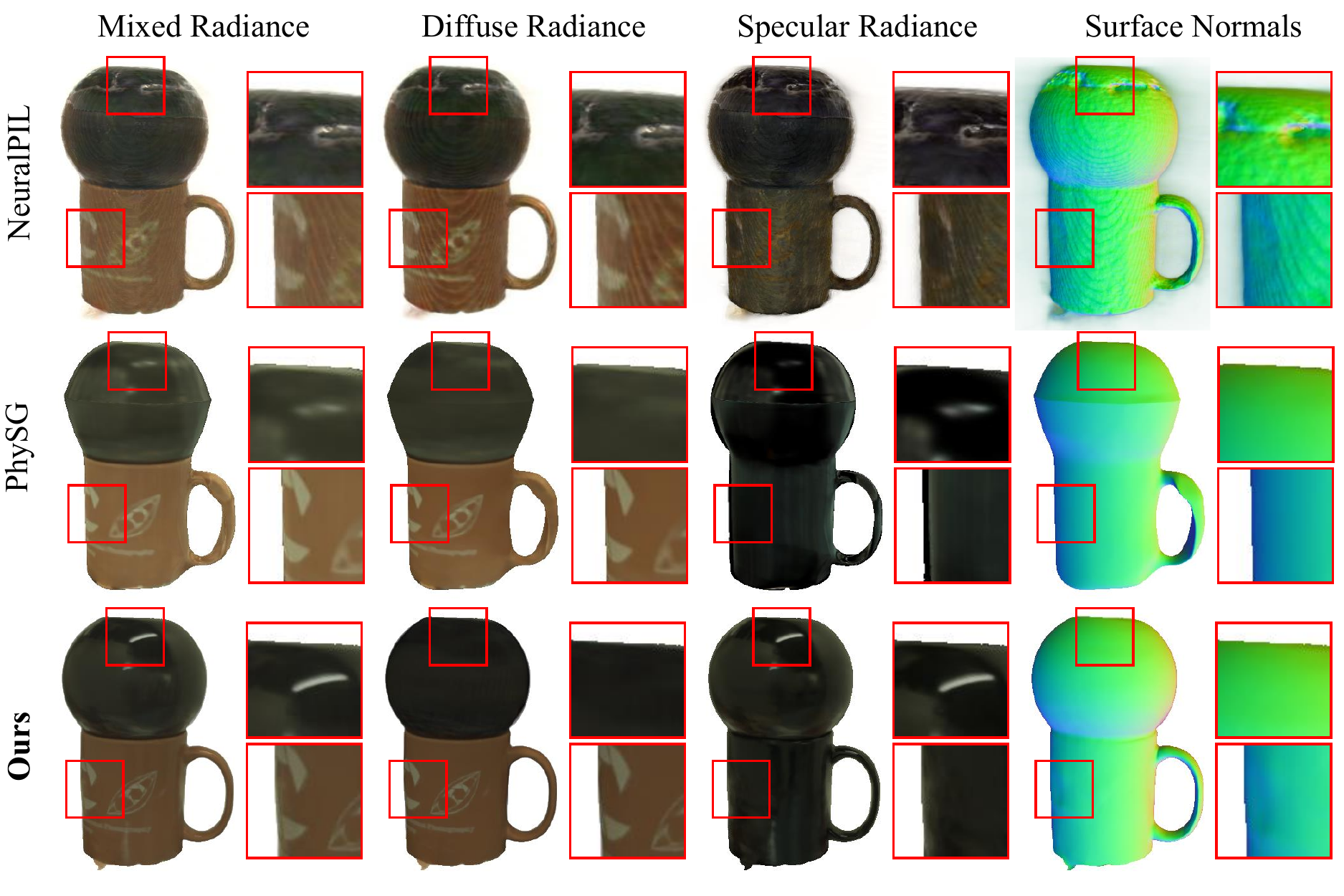}
\caption{\textbf{Reflectance separation and surface normal estimation on real dataset} The decomposition using PhySG and NeuralPIL on intensity-only images has artifacts such as the specular highlights bleeding into the diffuse component and surface normals. PANDORA on polarized images produces accurate diffuse radiance, models the sharp specular reflections and reconstructs precise surface geometry. }
\label{fig:real_radiance}
\end{figure}

\para{3D Reconstruction}
The polarization cues directly depend on the surface normals (\S \ref{sec:diff_spec_pol}). Thus, inclusion of polarization cues, enhances multi-view 3D reconstruction. PANDORA reconstructs cleaner and more accurate surfaces such as jaw of the bust in Fig. \ref{fig:sim_radiance} and the glass ball in Fig. \ref{fig:real_radiance}. In table \ref{table:metrics}, we show that the mesh reconstructed by PANDORA has much lower Hausdorff distance with the ground truth mesh as compared to state-of-the-art. PANDORA also estimates more accurate surface normals as evaluated on a held-out test set.  

\para{Diffuse-Specular Separation}
The inherent ambiguity in separating diffuse and specular radiance components from intensity-only measurements leads to artefacts in existing techniques. For example, the black sphere in diffuse radiance reconstructed by NeuralPIL and PhySG contain faint specular highlights. Difference in polarization of diffuse and specular components enables PANDORA to obtain more accurate separation along with better combined radiance Fig. \ref{fig:sim_radiance},\ref{fig:real_radiance}. In table \ref{table:metrics}, we show that PANDORA consistent outperforms state-of-the-art in peak signal-to-noise ratio (PSNR) and the structural similarity index measure (SSIM) of diffuse, specular and the net radiance images. We also provide the video of multi-view renderings from these diffuse, specular and mixed radiance fields that highlight the high quality of PANDORA's separation.



\begin{figure}[t!]
\begin{center}
      \includegraphics[width=\linewidth]{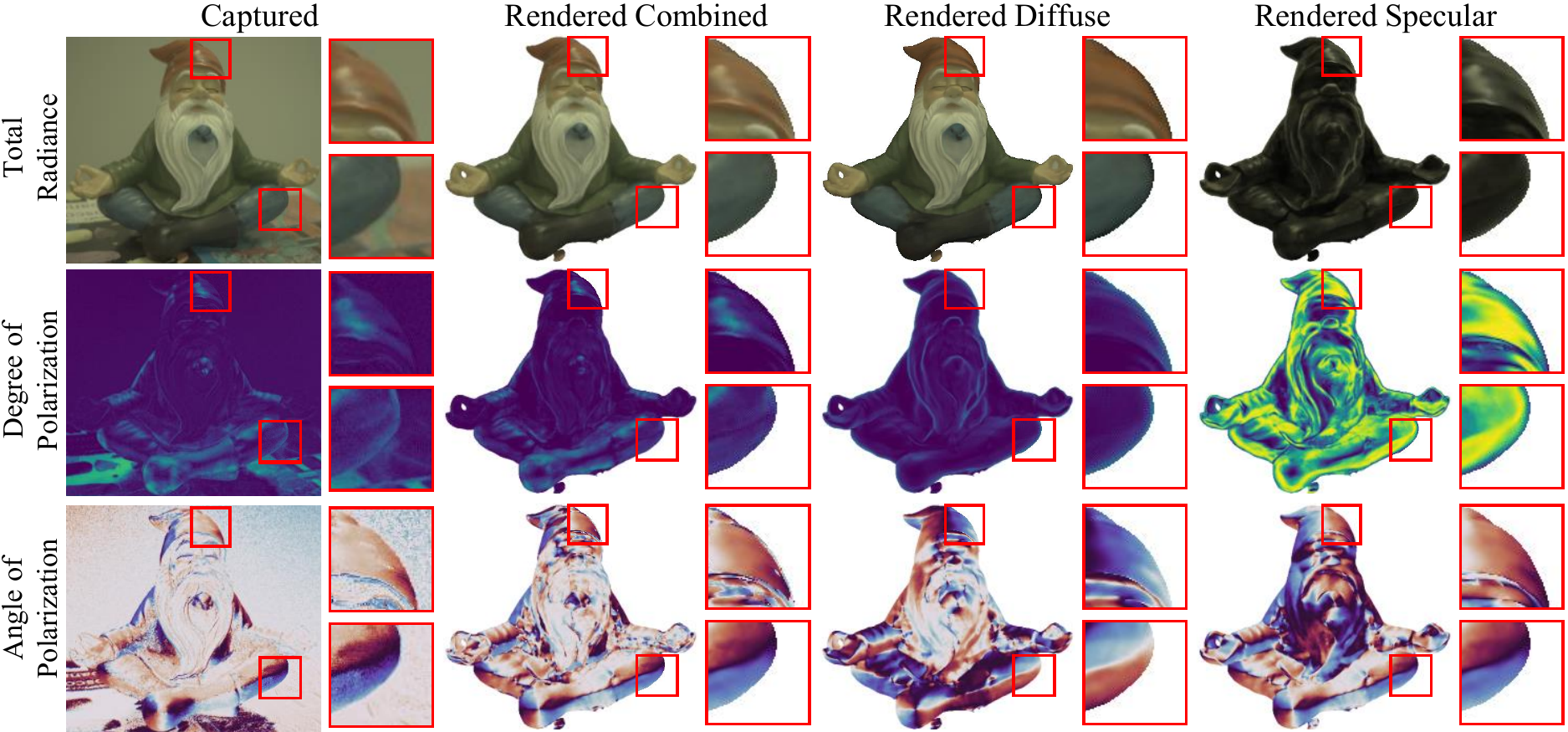}
\end{center}
\caption{\textbf{Polarimetric diffuse-specular separation on real-world objects}. PANDORA can separate out diffuse and specular polarimetric properties from captured polarized images. As expected, rendered specular component has higher degreee of polarization and polarization angle is orthogonal to the diffuse component. Polarization properties of the net rendered image match to that of the captured image.}
\label{fig:real_diff_spec}
\end{figure}
Apart from the radiance, PANDORA can also separate the polarization properties of the object's diffuse and specular components (Fig. \ref{fig:real_diff_spec}). Here, we see predicted cues match with our physical intuition: the AoP is orthogonal for the diffuse and specular components, while the DoP is higher for the specular component.

\para{Illumination estimation}
In addition to reflectance separation, our method can also estimate the illumination incident on the object. The rendered bust in Fig. \ref{fig:sim_radiance} has blurry specular highlights that make illumination estimation challenging.  We observe that NeuralPIL fails to estimate the correct lighting. PhySG employs spherical Gaussians that result in blurrier and more sparse reconstruction. PANDORA provides the best reconstruction with sharper walls and edges of the window.

 
\begin{figure}
\centering
\includegraphics[width=\linewidth]{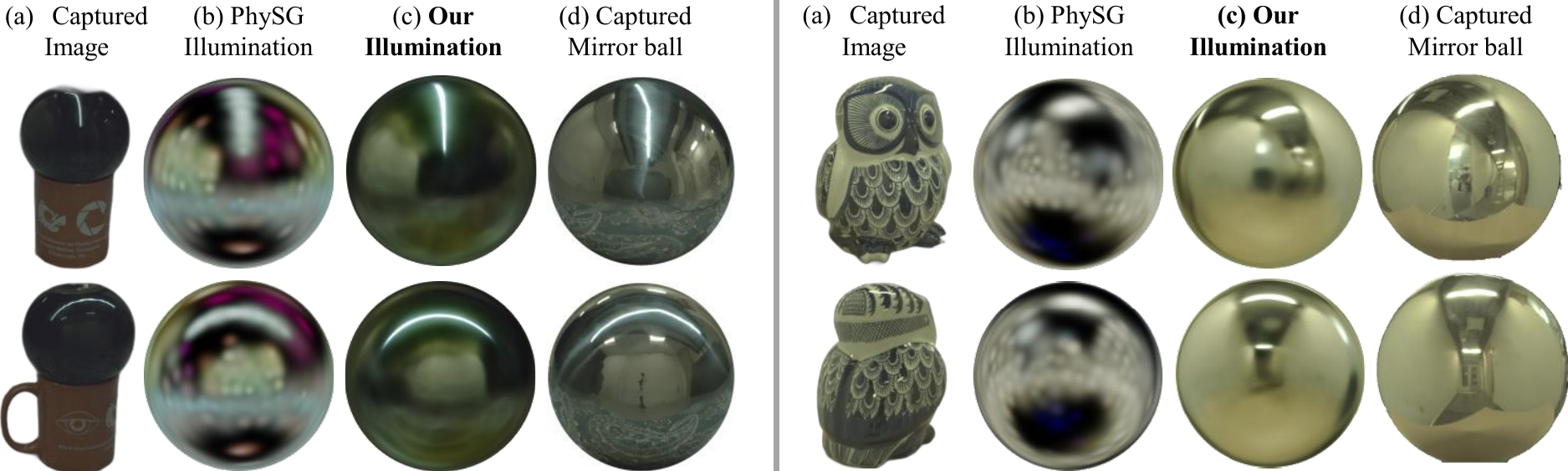}
\caption{\textbf{Incident illumination estimated from real object} We visualize the illumination estimated on a mirror ball viewed from two train viewpoints. We also capture a mirror ball placed at a similar viewpoints. PhySG models the illumination using spherical Gaussians and leads to blurrier reconstruction with artefacts. PANDORA's estimation has higher sharpness and accuracy.}
\label{fig:real_illum}
\end{figure}
Similarly, we also perform illumination estimation on real-world data (Fig. \ref{fig:real_illum}). We show results on data captured in two different environments. Fig. \ref{fig:real_illum}(left) is captured on a lab table with a long bright linear LED with dim ambient light. Fig. \ref{fig:real_illum}(right) is captured in a office hallway with many small tube-lights and bright walls. PhySG reconstruction is blurrier especially for the walls and comprises of color artifacts. PANDORA can recover high quality illumination that accurately matches the ground truth illumination as captured by replacing the object with a chrome ball.     

\subsection{Additional applications}
    The decomposed radiance field from PANDORA enables not only to render the object under novel views but also to change the object's appearance under novel views by altering the separated diffuse and specular radiance fields. We demonstrate this application under Fig. \ref{fig:teaser}(c). We perform polarimetric rendering from the learned PANDORA model under a new viewpoint. The rendered polarization (Fig. \ref{fig:teaser}(c) bottom left) is consistent with the captured polarization. As PANDORA decomposes radiances, we can alter the diffuse component without affecting the specular reflections. For example, we assign pink albedo to the object by removing the G component of radiance without altering the color of the specularities in (Fig. \ref{fig:teaser}(c) top left). To make the object look metallic, we render only the specular component with the Fresnel reflectance $R^+$ set to 1 (Fig \ref{fig:teaser}(c) top right). To obtain rougher appearance  (Fig \ref{fig:teaser}(c) bottom right), we multiply the roughness parameter with a factor 3 before passing to the IllumNet. Please refer to \webpage{project webpage} for multi-view renderings of the separated reflectance fields and the changed appearance. 

\begin{figure}[!ht]
    \centering
    \includegraphics[width=\linewidth]{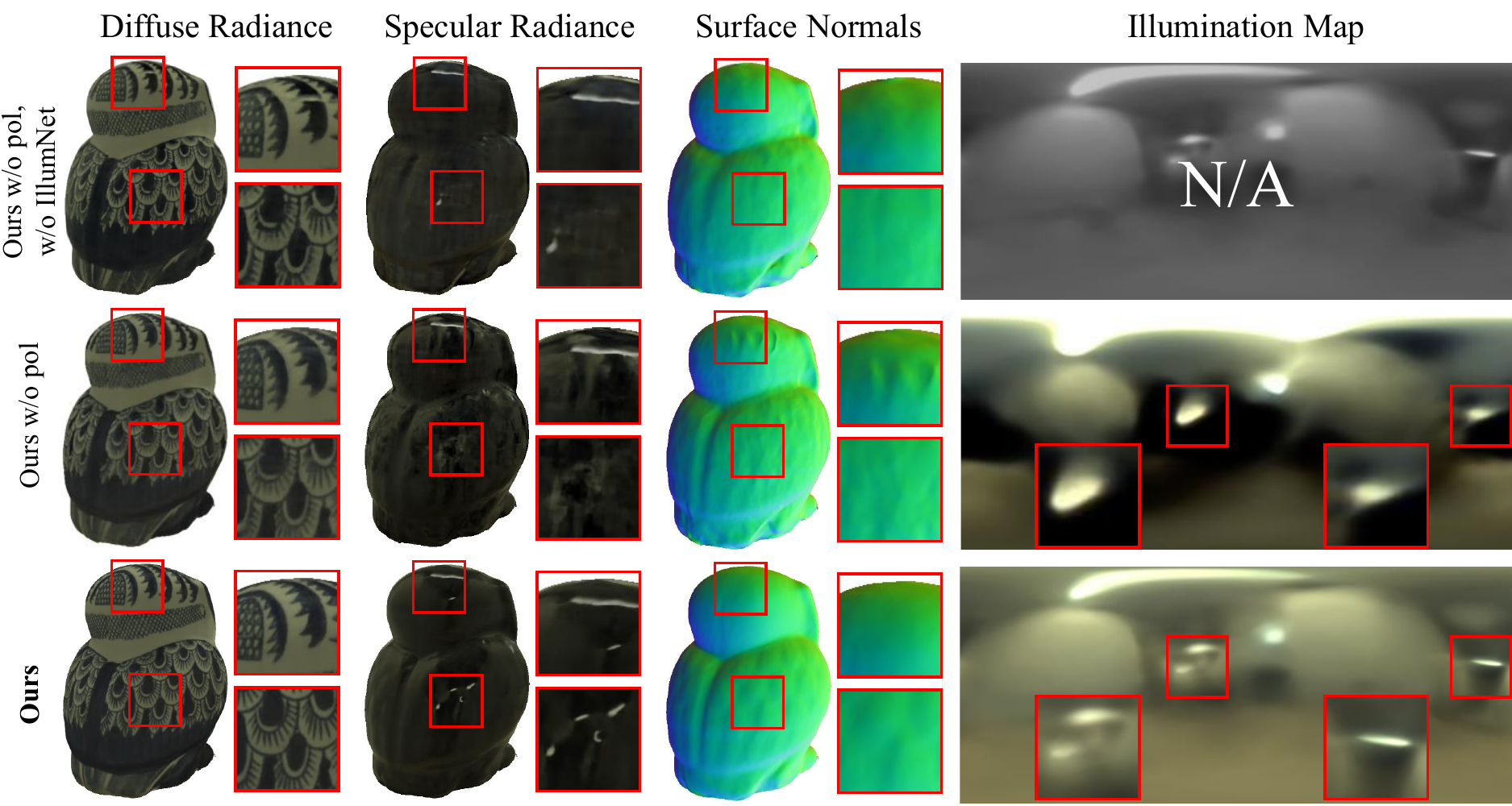}
    \caption{\textbf{Ablation study: Role of polarization and IllumNet} We devise two ablation experiments by running PANDORA on intensity-only images without IllumNet(top row) and with IllumNet(middle row). Without polarization and correct illumination modelling, there are texture artefacts in specular and surface normals due to the ambiguity in texture decomposition. Polarimetric cues and IllumNet help PANDORA in resolving such ambiguities resulting in finer quality reconstructions with sharper diffuse texture, cleaner surface normals and accurate lighting estimation.}
    \label{fig:real_ablation}
\end{figure}
\subsection{Ablation study: Role of polarization and IllumNet} 
Polarization and illumination modelling are key aspects of PANDORA. Here we analyse the role of these components by devising the following experiments
\begin{tight_enumerate}
    \item Ours w\textbackslash o IllumNet w\textbackslash o pol: We set Stokes loss weightage factor $w_s$ (Eq. \ref{eq:L_stokes}) as 0 to constrain PANDORA to just use $S_o$ component, i.e., intensities for radiance decomposition. Also, instead of modelling illumination and roughness with neural networks, we directly model the specular radiance with a neural network same as the conventional RadianceNet in VolSDF.
    \item Ours w\textbackslash o pol: We set $w_s$ as 0. But keep the IllumNet. So, this model has the same architecture as PANDORA. But it is trained on only the intensity.
\end{tight_enumerate}
We then train these two models and PANDORA on the same data with the same training scheme. As shown in Fig. \ref{fig:real_ablation}, inclusion of the illumination modelling and polarization information significantly improves PANDORA's performance. The model without IllumNet and polarization, exhibits strong artefacts of specular highlights in the diffuse and fails to capture the smaller specularites. Removing just polarization leads to worse illumination estimation, bleeding of diffuse into the specular and texture artefacts in the normals. Equipped with polarization information and correct illumination modelling, PANDORA outputs sharper diffuse texture, accurately handles the small specularities and even captures the subtle bumps on the object's back.

\section{Conclusion and Discussion}
We have proposed PANDORA a novel neural inverse rendering algorithm that achieves state-of-the-art performance in reflectance separation and illumination estimation. PANDORA achieves this by using polarimetric cues and an SDF-based implicit surface representation. We have demonstrated the success of our approach on both simulated data that was generated by a physics based renderer, and real-world data captured with a polarization camera. Finally, we compared against similar approaches and demonstrated superior surface geometry reconstruction and illumination estimation. We believe PANDORA would pave the way for exciting ideas in the space of polarimetric and neural inverse rendering.

\clearpage


\bibliography{PANDORA_bib}
\appendix

\section{Forward Model Derivation}
\label{sec:suppl_derivation}
In this section, we elaborate on the derivation of exitant Stokes vector as a function of diffuse and specular radiance as described in Eq. 3 of the main manuscript.
\paragraph{Diffuse Component}
In Eq. 2, we decompose the outgoing Stokes vector into diffuse and specular components. First we focus on the diffuse component. From the definition of $H_d$ for pBRDF model \cite{baek2018simultaneous} and the illumination Stokes vector defined in eq. 1, we obtain
\begin{align}
    H_d \cdot S_i = \rho(\bn\cdot\bi)L_iT_i^+T_i^-
\begin{bmatrix}
T^+_o \\
T_o^-\alpha^{}_{o} \\
-T_o^-\delta^{}_{o} \\
0 \\
\end{bmatrix}\;,
\end{align}
where $\rho$ is the diffuse albedo, $\bn$ is the surface normal and $\bi$ is the incident illumination direction. 
 With $\phi_n$ denoting the exitant azimuth angle w.r.t. the surface normal, we define $\alpha^{}_{o}$ and $\delta^{}_{o}$ as 
\begin{align}
\begin{split}
    \alpha^{}_{o} &= \cos\left(2\phi_\mathrm{n}\right)\\
    \delta^{}_{o} &= \sin\left(2\phi_\mathrm{n}\right)
\end{split} 
\end{align}
We denote the term  $\rho(\bn\cdot\bi)L_iT_i^+T^+$ as the diffuse intensity $L_D$. The term $H_d \cdot S_i$ is independent of the viewing direction. Thus we obtain the first component of Eq.3
\begin{align}
   \int_{\Omega}{H_d \cdot S_i(\mathbf{x},\omega_i)d\omega} =  
     L_d    \begin{bmatrix}
        1 \\
        T^-_o/T^+_o \cos(2\phi_n) \\
        -T^-_o/T^+_o \sin(2\phi_n)\\
\end{bmatrix}
\end{align}
\paragraph{Specular Component}
The specular exitant Stokes vector is obtained by substitution of  $H_s$ as defined in the pBRDF model \cite{baek2018simultaneous} and $S_i$ from eq. 1.
\begin{align}
   H_s \cdot S_i = L_i\dfrac{k_sDG}{4(\bn\cdot\bo)}
   \begin{bmatrix}
         R^+\\
         R^-\chi^{}_o \\
         R^-\gamma^{}_o \\
   \end{bmatrix}.
   \label{eq:s_s}
\end{align}
where $k_s$ is the specular coefficient, $\bo$ is the exitant direction, $D$ is the microfacet distribution and $G$ is the microfacet shadowing term.
With $\varphi_h$ and $\varphi_h$ denoting the incident and exitant azimuth angle w.r.t. the half angle $\mathbf{h}$ respectively, we define $\chi^{}_{o}$ and $\gamma^{}_{o}$ as 
\begin{align}
\begin{split}
    \chi^{}_{h} &= \sin\left(2\varphi_h\right)\\
    \gamma^{}_{h} &= \cos\left(2\varphi_h\right)
\end{split}
\end{align}
We denote $f_s = \dfrac{k_sDGR+}{4(\bn\cdot\bo)}$. Theoretically $\chi$ and $\gamma$, depend on the half angles and not the geometric surface normals of the object. In practice, we observe that for realistic values of the roughness, $\chi$ and $\gamma$ do not significantly deviate from the value obtained using surface normals instead of the half angle,i.e. $\chi^{}_{h} \approx \sin\left(2\phi_h\right)$ and $\gamma^{}_h \approx \cos\left(2\phi_n\right)$. As a result,
\begin{align}
       \int{(H_s \cdot S_i) di} = L_i\dfrac{k_sDG}{4(\bn\cdot\bo)}
   \begin{bmatrix}
         R^+\\
         R^-\chi^{}_o \\
         R^-\gamma^{}_o \\
   \end{bmatrix} \int{f_s L_i di}.
   \label{eq:int_s_s}
\end{align}
We denote $R^+\int{f_s L_i di}$ as the specular radiance $L_s$ and obtain the specular component of the output Stokes vector
\section{Implementation Details}
\begin{figure}
    \centering
    \includegraphics[width=0.8\textwidth]{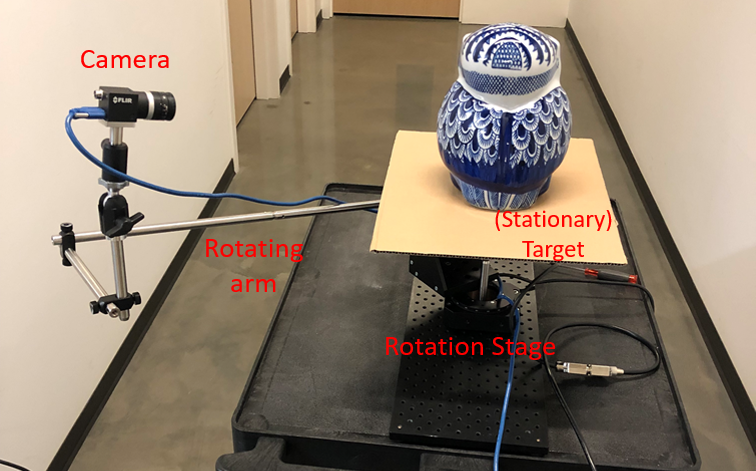}
    \caption{\textbf{Experimental Setup:} Above, is an image of our experimental setup. The target object is placed on the stationary section of a rotation stage, which is attached to an extended arm and the snapshot polarimetric camera. The camera capture polarimetric images from multiple angles under unstructured lighting while the target object remains still. }
    \label{fig:exp_setup}
\end{figure}

Real world data was captured with a Blackfly S USB3 camera with Sony IMX250MYR Polarization-RGB sensor \cite{SonyPol}. 35 images were captured for the Ball-Cup, Owl and Gnome objects under different lighting conditions as described in Table 2. The camera was placed along multiple angles distributed roughly equally along a circle around the target object using a portable setup as shown in Fig. \ref{fig:exp_setup}. To capture the ground truth illumination map as shown in Fig. \ref{fig:suppl_real_radiance} last row, we use the same setup and flip the camera so that it points outside instead of the scene. Fish eye lens is used to increase the field of view and multi-view images are captured and stitched together to obtain the ground truth illumination map. 

\begin{figure*}
\begin{center}
   \includegraphics[width=0.8\linewidth]{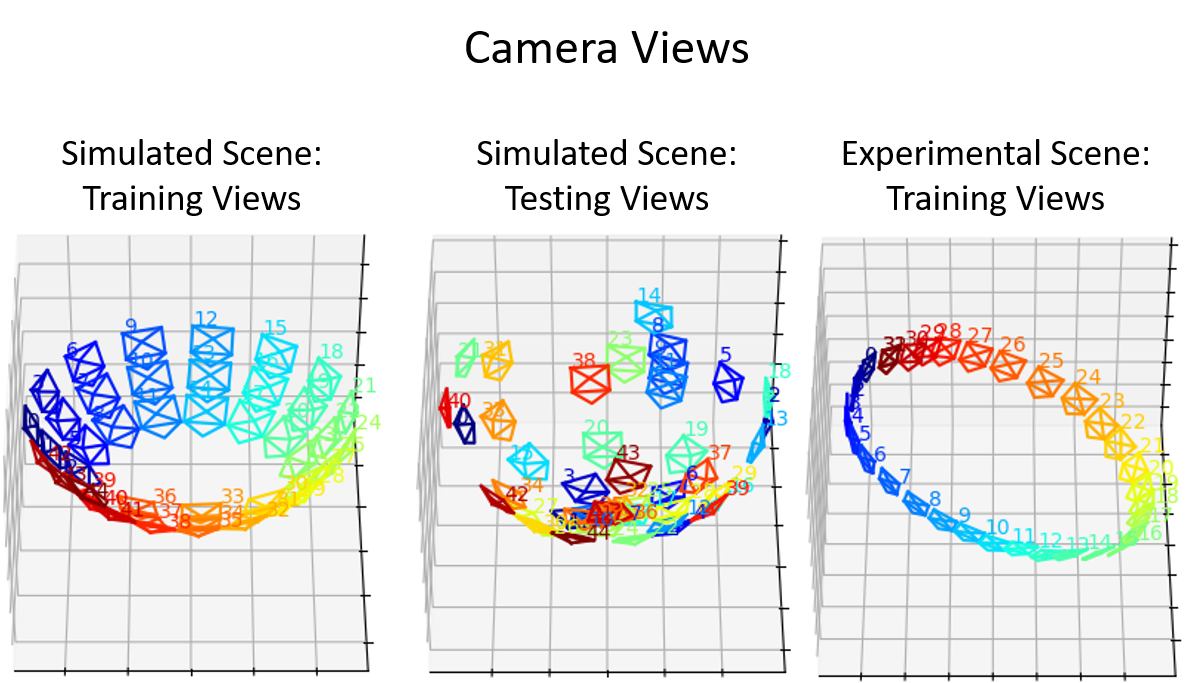}
\end{center}
  \caption{\textbf{Camera Views} Above, we show the camera positions for both the simulated and experimentally captured data.}
  \label{fig:cam_views}
\end{figure*}

\paragraph{Rendering data generation} Simulated data is generated using the Mitsuba2 renderer \cite{NimierDavidVicini2019Mitsuba2}. In Mitsuba2, we are able to set the material properties, camera angles, illumination, and imaging modality (polarized or unpolarized). We use a brdf that possesses equally weighted diffuse and dielectric (specular) components. We use 45 camera views distributed over all azimuth angles, and range from 25 to 50 degrees in elevation. Our two ground truth targets were a standard sphere and a bust shape obtained from \cite{augustusmodel}. The camera views are shown in Fig. \ref{fig:cam_views}.

\paragraph{Training details}
\label{sec:suppl_details}

All training and testing was conducted on a server containing Nvidia 2080 Ti's. As stated in our main body, our DiffNet, MaskNet, RoughNet, and IllumNet were standard MLPs with 4 layers and a width of 512. Our SDFNet was an 8-layer MLP with a width of 256 and a single skip connection in the 4-th layer. Our training procedure uses several hyperparameters. The most relevant parameters include the weightage of the stokes vector loss, the weight of the mask network loss, the number of warm up iterations (before the stokes vector and specular components are estimated), and the total number of iterations. For real-world data we use 1000 warm-up iterations and 100,000 total iterations, while for simulated data we use 1500 warm-up iterations and 50,000 total iterations. We empirically found that a mask loss weightage and stokes loss weightage of 1.0 and 0.1, respectively, produced high-quality results. The diffuse and mask networks used a sigmoid activation function, while the specular and roughness networks used a softplus activation function to avoid vanishing gradients. Finally, for our SDFNet, MaskNet, RoughNet, and IllumNet, we used the frequency embeddings described by Mildenhall et al \cite{mildenhall2020}. The frequencies of the embeddings were sampled in log-space from $2^0-2^6$ for the SDFNet and from $2^0-2^{10}$ for the MaskNet, RoughNet, and DiffNet. The integrated directional embeddings were used to embed the directional coordinates for the IllumNet, as described in more detail in the subsequent section.

\paragraph{Illumination Network Design}
\begin{figure}
    \centering
    \includegraphics[width=1.0\linewidth]{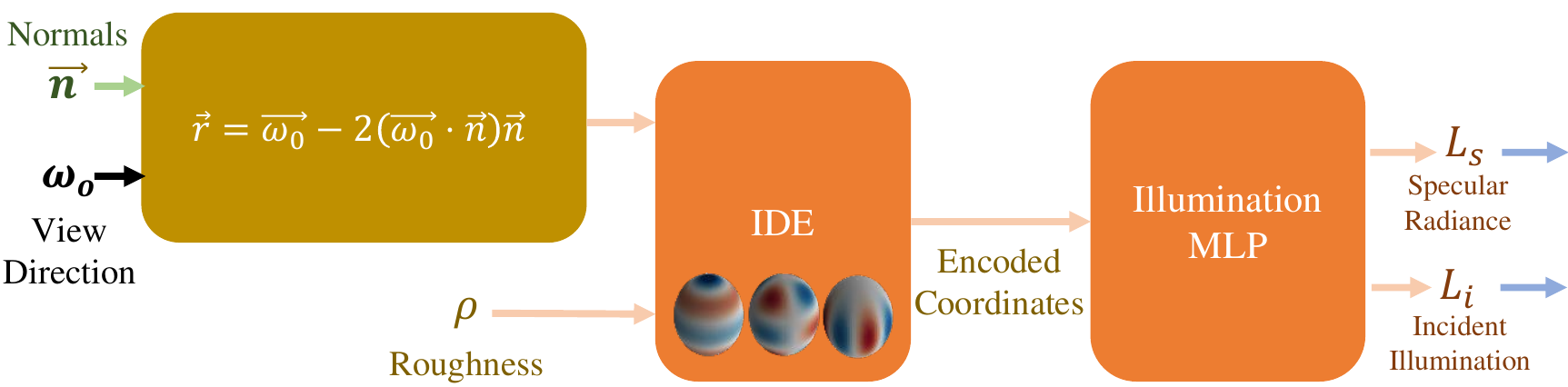}
    \caption{\textbf{Illumination Network Design}: The illumination network accepts the reflected direction vector and the predicted surface roughness as input. The reflected direction is calculated from the surface normal and viewing direction as shown above. The roughness and direction vector are encoded by the IDE before it is passed to the MLP which generates the predicted illumination and radiance based on fresnel reflectance.}
    \label{fig:suppl_illum_Net}
\end{figure}
The illumination network is responsible for calculating the incident illumination (the environment map) and the specular radiance, which is derived from Fresnel reflectance. To do this, the network accepts the reflected direction and the roughness as input. The roughness parameter is estimated by a separate network, while the reflected direction can be calculated from the predicted surface normals (using the geometry network) and the input viewing direction. Both inputs must be encoded through the IDE to help estimate the high frequency information and incorporate the effects of the roughness parameter, i.e. increase the blurring of the predicted illumination as the roughness gets larger. For the input to our IllumNet, we used degree $L\in\{1,2,4\}$ spherical harmonics with order $m\in[-L,L]$ for the IDE's.
\section{Additional Results}
\label{sec:suppl_additional}
\begin{center}
\centering
\small
\begin{tabular}{|p{0.5 cm} c |p{3cm}|cc|cc|cc|c|}
 \hline
    \multicolumn{2}{|c|}{\multirow{3}{*}{Scene}}& {\multirow{3}{*}{Approach}} & \multicolumn{2}{c|}{Diffuse}& \multicolumn{2}{c|}{Specular}& \multicolumn{2}{c|}{Mixed} & Normals\\
&&&  PSNR &  SSIM & PSNR & SSIM & PSNR &  SSIM  & MAE\\
&&&  $\uparrow$ (dB) &  $\uparrow$ &  $\uparrow$ (dB) & $\uparrow$ & $\uparrow$ (dB) &  $\uparrow$  &$\downarrow$ (\degree) \\
 \hline
{\multirow{3}{*}{\rotatebox[origin=c]{90}{\hspace{0cm}Bust}}} &  \multirow{3}{*}{\includegraphics[width=1.cm]{out_imgs/gt_masked_s0/aug_v33/gt_000_mixed.png}} & NeuralPIL&23.90&0.87&18.04&0.87&26.71&0.87&15.36\\
&&PhySG &22.64 & 0.94 & 23.00 & 0.94 & 19.94 & 0.72 & 9.81\\
&&Ours no pol no Illum & 28.29 & 0.968 & 21.13 & 0.906& 22.29 & 0.951 & 7.89\\
&&Ours no pol& 25.78 & 0.956 & 18.23 & 0.856 & 22.50 & 0.927 & 4.83\\
&&Ours & 29.53 & 0.973 & 23.63 & 0.912 & 25.97 & 0.951 & 1.95\\
 \hline
{\multirow{3}{*}{\rotatebox[origin=c]{90}{\hspace{0cm}Sphere}}}&   \multirow{3}{*}{\includegraphics[width=1.cm]{out_imgs/gt_masked_s0/christmas_sphere_v11/gt_000_mixed.png}}&NeuralPIL &13.09&0.55&12.92&0.55&20.04&0.66&38.73\\
&&PhySG& 21.76 & 0.76 & 18.90 & 0.76 & 17.93& 0.70& 8.42\\
&&Ours no pol no Illum & 20.65 & 0.76 & 16.23 & 0.76 & 17.11 & 0.72 & 1.91\\
&&Ours no pol& 22.20 & 0.83 & 21.30 & 0.87 & 20.87 & 0.82 & 1.92\\
&&Ours & 24.29 & 0.84 & 21.29 & 0.88 & 21.29 & 0.83 & 1.04\\
 \hline
\end{tabular}
\captionof{table}{\textbf{Quantiative evaluation on rendered scenes} We evaluate PANDORA with state-of-the-art and ablation methods on held-out testsets of 45 images for two rendered scenes. We report the peak average signal-to-noise ratio (PSNR) and structured similarity (SSIM) of diffuse, specular and net radiance and mean angular error (MAE) of surface normals. PANDORA consistently outperforms state-of-the-art in radiance separation and geometry estimation.} 
\label{table:sim_suppl_metrics}
\end{center}
\begin{figure*}
\centering
\includegraphics[width=0.9\linewidth]{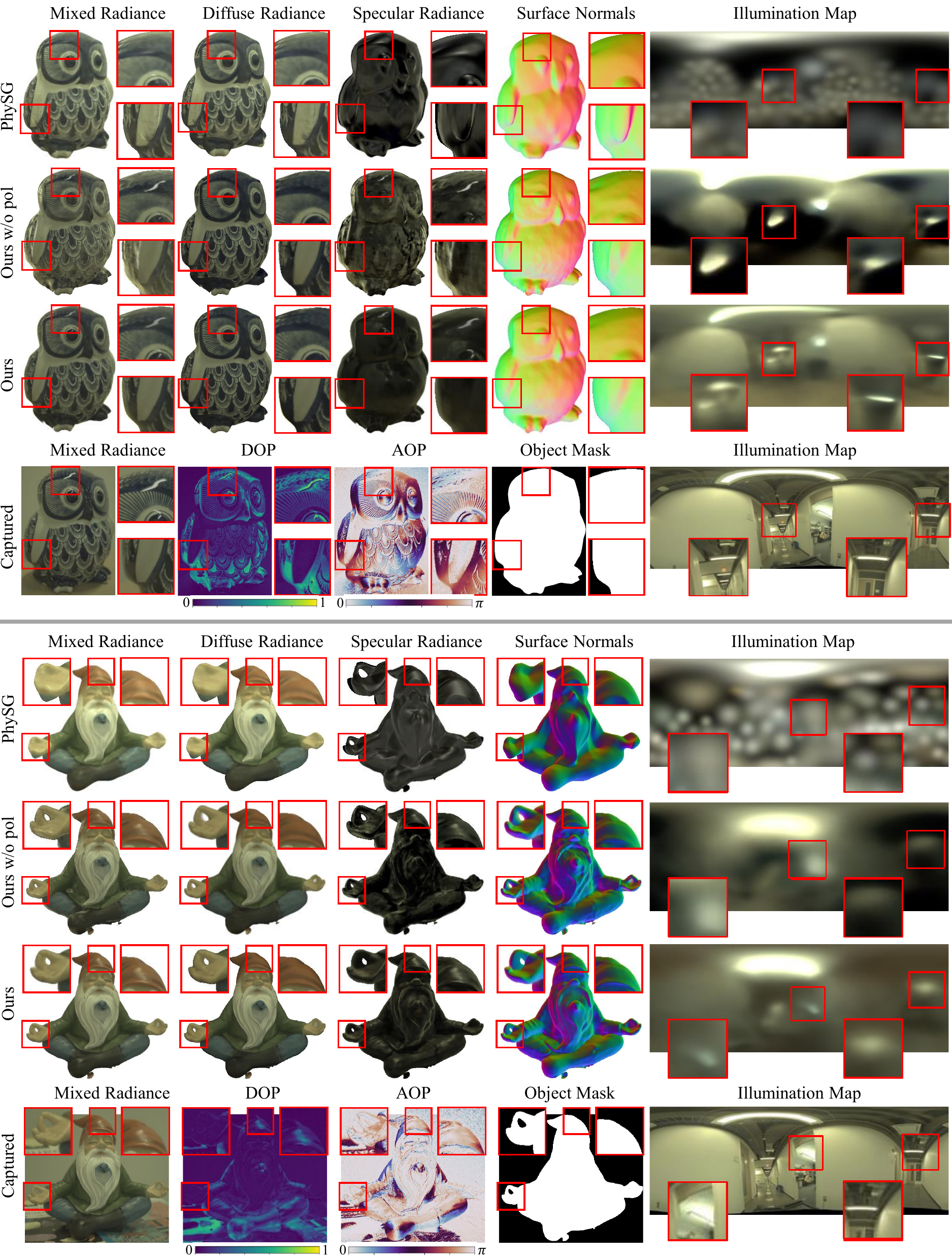}
\caption{\textbf{Reflectance separation,surface normal reconstruction and illumination estimatoin on real dataset} PANDORA captures high frequency details in the surface normals and accurately models the specular highlights. Please view the \webpage{project webpage} for multi-view renderings of the same.}
\label{fig:suppl_real_radiance}
\end{figure*}
\begin{figure}
\centering
\small
\begin{minipage}{0.29\textwidth}
\includegraphics[width=\textwidth]{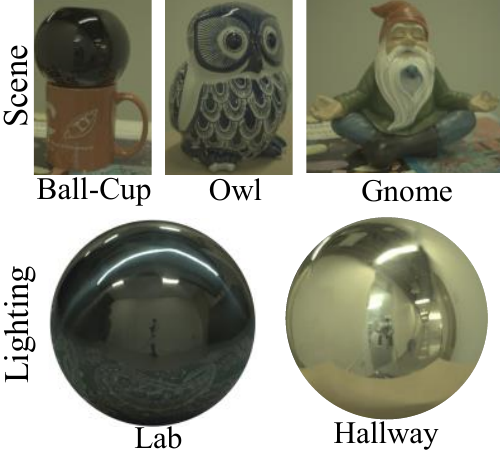}
\end{minipage}
\begin{minipage}{0.7\textwidth}
\begin{tabular}{|l |m{1.6cm}|cc|cc|cc|}
 \hline
    \multicolumn{1}{|c|}{\multirow{3}{*}{Scene}}& {\multirow{3}{*}{Lighting}} & \multicolumn{2}{c|}{PhySG}& \multicolumn{2}{c|}{Ours w\textbackslash o pol}& \multicolumn{2}{c|}{Ours} \\
&&  PSNR &  SSIM & PSNR & SSIM & PSNR & SSIM \\
&&  $\uparrow$ (dB) &  $\uparrow$ &  $\uparrow$ (dB) & $\uparrow$ & $\uparrow$ (dB) &  $\uparrow$ \\
 \hline
 \multirow{1}{*}{Owl} & Hallway& 27.68& 0.953& 27.67& 0.940& 30.37& 0.960\\
 \hline
 \multirow{1}{*}{Gnome}&Hallway & 30.31&0.986 &28.42 & 0.984 & 29.15&0.984 \\
 \hline
 \multirow{1}{*}{Ball-cup}&Hallway &19.46&0.920& 27.99& 0.980& 28.12& 0.981\\
 \hline
 \multirow{1}{*}{Ball-cup}& Lab& 14.00& 0.950 & 23.52& 0.953& 26.92& 0.970\\
 \hline
\end{tabular}
\end{minipage}
\label{table:real_metrics}
\captionof{table}{\textbf{Quantiative evaluation on real scenes} We report the average PSNR and SSIM of the rendered intensity image over the training set for objects with different material properties and under different lighting conditions. PANDORA consistently outperforms PhySG and the ablation model that is devoid of the polarimetric cues. } 
\end{figure}
\begin{figure}[t!]
\centering
\includegraphics[width=\linewidth]{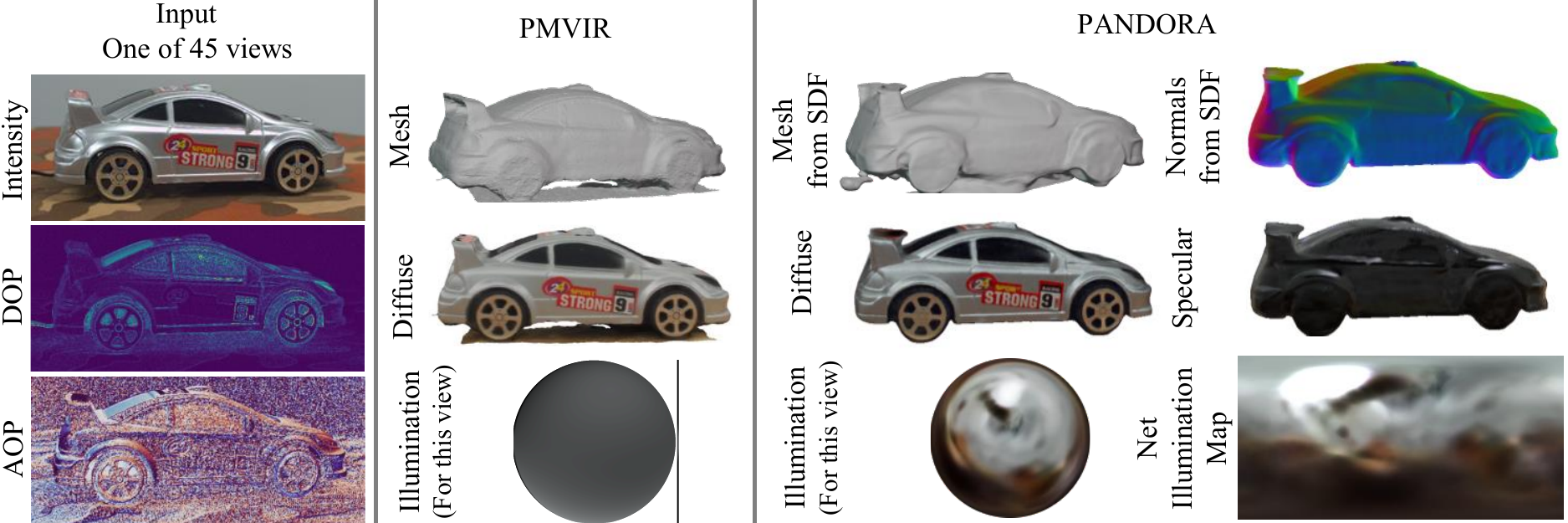}
\caption{\textbf{Comparison with prior mesh-based polarimetric inverse rendering on real data} Utilizing similar multi-view snapshot polarimetric data as ours, PMVIR \cite{Zhao2020} recovers 3D mesh, diffuse color for mesh vertex and lighting based on diffuse shading. Neural implicit representations enable PANDORA to extract more from the same captured data. PANDORA learns the continous signed distance field from which mesh and surface normals can be extracted. Apart from the diffuse color, PANDORA also outputs the specular radiance. Illumination estimated from PANDORA features sharp light source and the orange floor that better explain the captured data.}
\label{fig:real_pmvir}
\end{figure}
    In Fig. \ref{fig:suppl_real_radiance}, we show additional qualitative comparisons with state-of-the-art inverse rendering technique, PhySG \cite{physg2020}, and ablation model run on intensity-only images. In Fig. \ref{fig:real_pmvir}, we highlight the advantages of PANDORA over existing mesh optimization-based polarimetric inverse rendering technique, PMVIR \cite{Zhao2020}. We also report additional quantitative metrics on simulated and real data in Table \ref{table:sim_suppl_metrics} and Table 2 respectively. Please refer to the \webpage{project webpage} for videos showcasing our multi-view renderings.
\section{Analysis}
\paragraph{Performance on out-of-distribution views}

\begin{figure}
    \centering
    \includegraphics[width=1.0\textwidth]{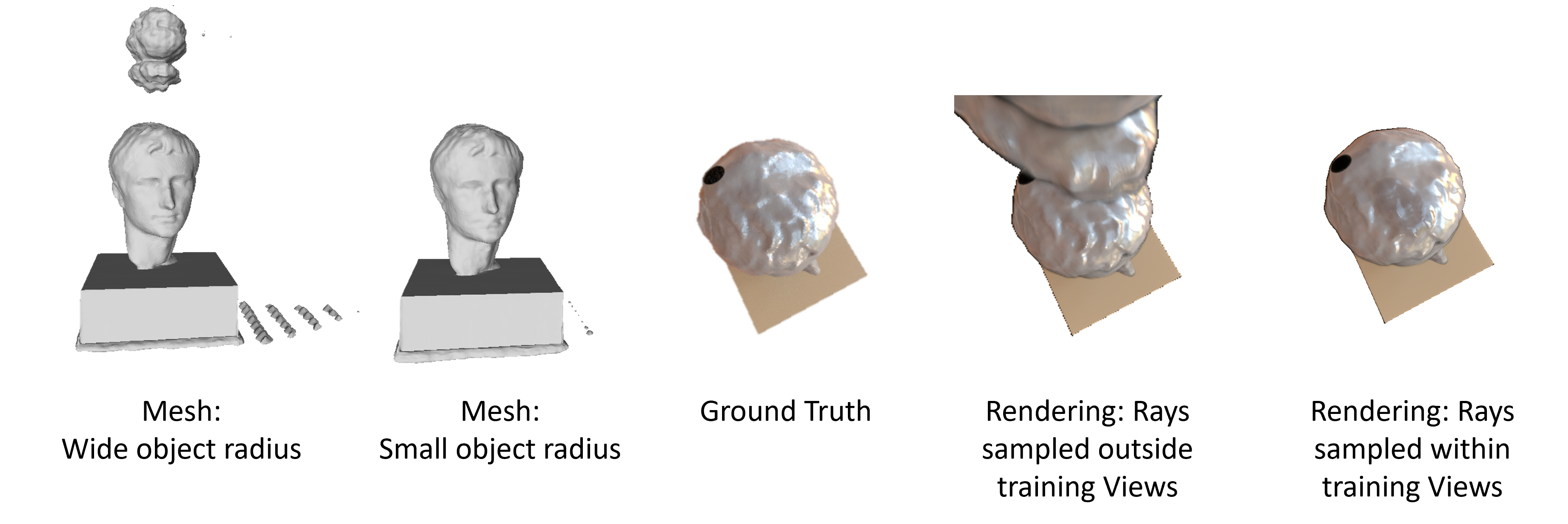}
    \caption{\textbf{Extrapolated Views Result:} We show the estimated mesh corresponding to regions that had lower sampled (panel 1). and higher sampled (panel 2) views. In addition, we show the resulting renderings when using more extrapolated rays (panel 4) versus without the extrapolated rays (panel 5). }
    \label{fig:extrap}
\end{figure}

As expected, for regions outside of the views in our training images, the estimation performs poorly. We see in Fig. \ref{fig:extrap} the network extrapolates a blob above the statue, in regions that are not heavily sampled during training. This affects our rendering when we sample rays in these regions (Fig. \ref{fig:extrap} panel 4). Finally, we see that by sampling rays only within a narrower region of interest, corresponding to locations with more training views, we obtain a correct estimate. We should note that in our main paper, the reported metrics do not account for this poor extrapolation as the images were rendered over a wider region of interest. So, the metrics were  affected by artefacts in some of the rendered images shown in Fig. \ref{fig:extrap} panel 4. Metrics reported in Table \ref{table:sim_suppl_metrics} are with images rendered over smaller region of interest and do not have this artefact.

\begin{figure}[t!]
\centering
\includegraphics[width=0.7\linewidth]{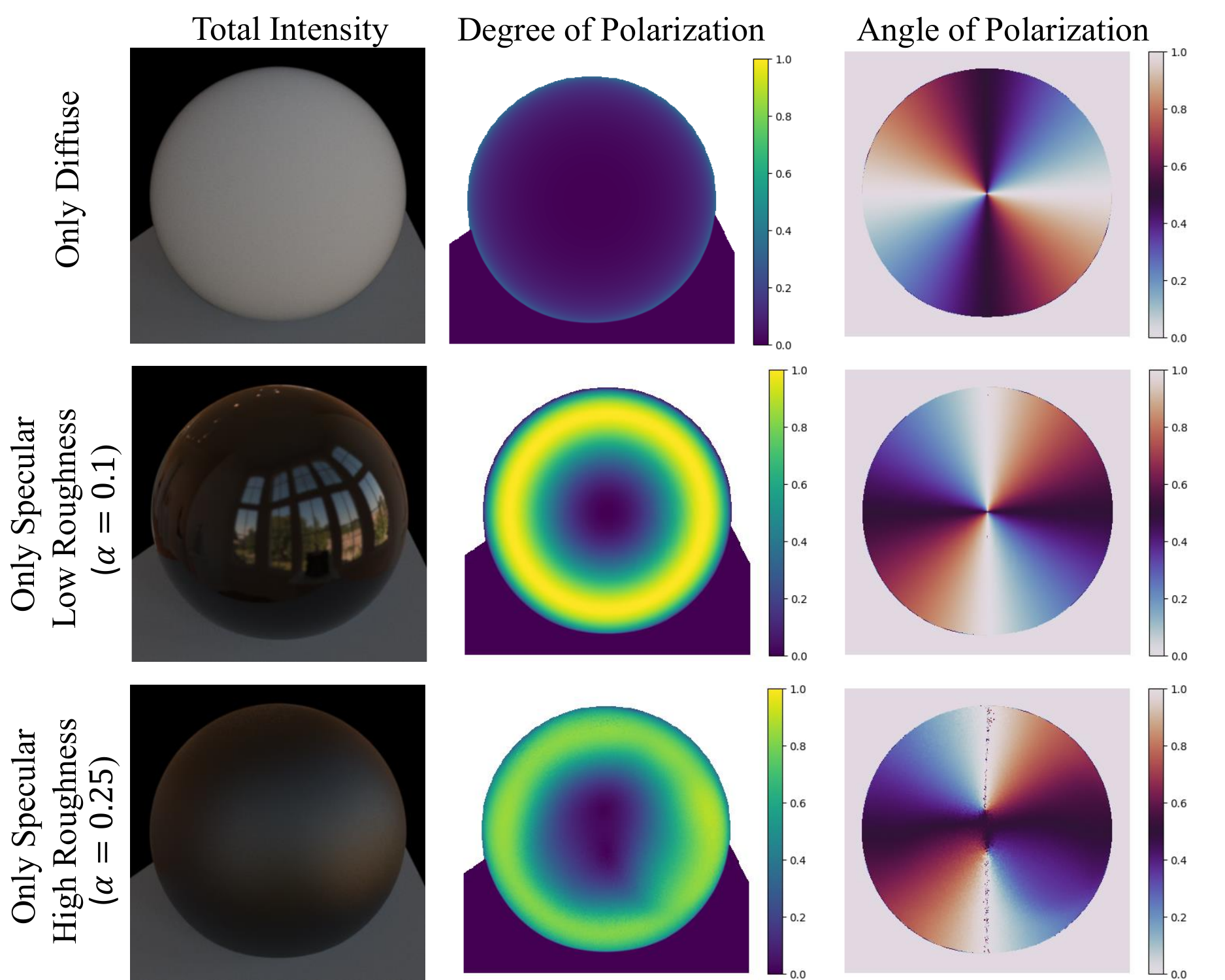}
\caption{\textbf{Effect of specular roughness on polarimetric cues} We render polarimetric cues for a sphere object using the pBRDF model in Mitsuba2 with varying material properties. The variation of polarimetric cues is less under the realistic range of roughness. Our insight that the specular polarization is orthogonal in angle and higher in degree than the diffuse polarization remains applicable on varying specular roughness to realistic values. $alpha$ denotes the roughness parameter of the Beckmann microfacet distribution.}\label{fig:sim_rough_pol_cues}
\end{figure}

\paragraph{Effect of roughness on illumination estimation}
\begin{figure}[ht!]
    \centering
    \includegraphics[width=1.0\linewidth]{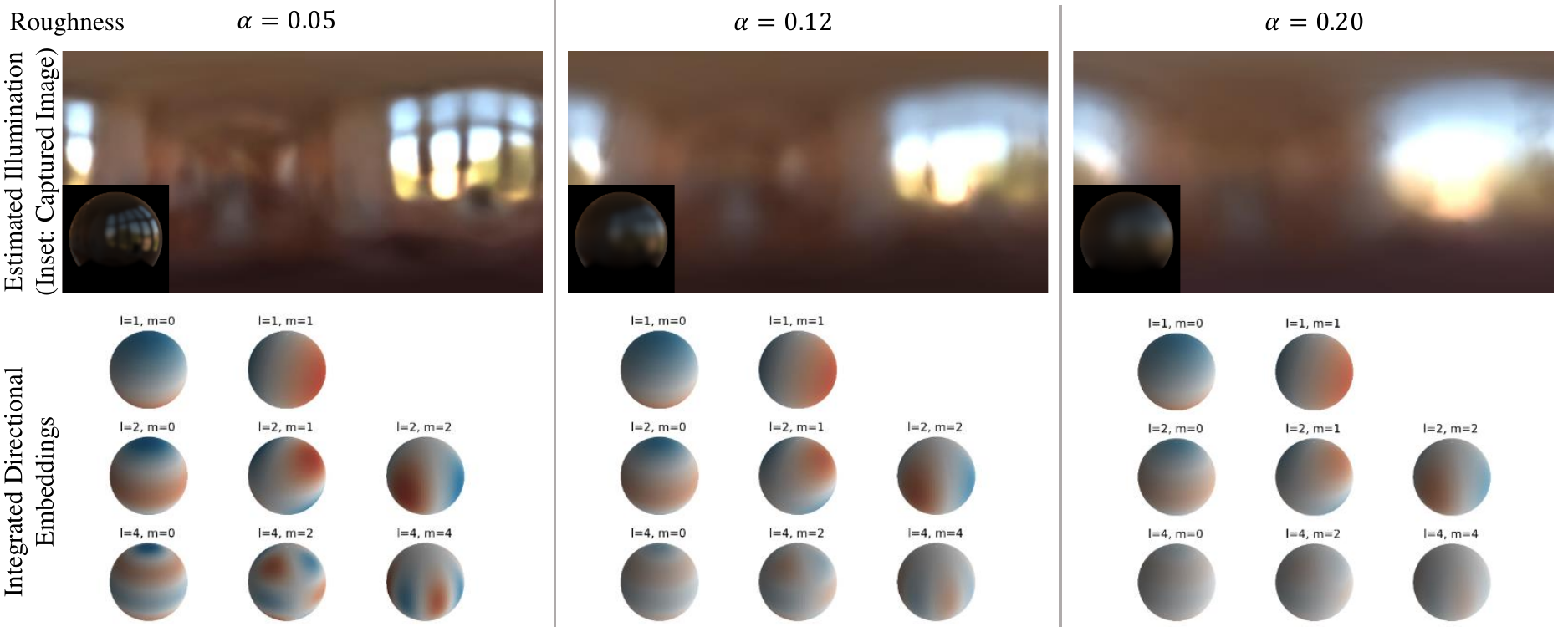}
    \caption{\textbf{Effect of roughness on illumination estimation}: Our illumination estimation accounts for the effects of surface roughness. As the roughness (parameterized by $\alpha$) increases, there is an increasing blurring effect on the estimated environment map. The inset images shows the corresponding ground truth specular reflection as the surface roughness increases. The right side shows the effect of the increasing roughness on the spherical harmonic IDE's.}
    \label{fig:suppl_roughness_test}
\end{figure}
Above, we show the effects of the surface roughness on the estimated illumination map. As the surface roughness ($\alpha$) increases, the associated, estimated environment map is increasingly blurred. The inset images show the ground truth specular reflection for each of the estimated environment maps. On the right-hand side, we show the associated spherical harmonic bases, which are used for the integrated directional encoding (IDE) \footnote{The IDE visualization was generated using the ReF-NeRF implementation, with help in implementing the spherical harmonics transform from \cite{yu2021plenoxels,se3transformer}} \cite{verbin2021ref}. Recall that the IDE is used to encode the directional coordinates, which are passed as input to the illumination MLP. Increasing roughness decreases the impact of the higher frequency spherical harmonic bases, as shown on the right. This helps to supervise the desired blurring effect because the high-frequency components reduced.
\paragraph{Effect of roughness on polarimetric cues.} In Fig. \ref{fig:sim_rough_pol_cues}, we show using renderings from Mitsuba that the variation of polarimetric cues on varying roughness is less and the polarization of specular component is always distinct from the diffuse polarization under different levels of roughness. 
\section{Limitations}
There are two main limitations to our current approach. Firstly, our method does not handle self-occlusions. This is more prominent in our simulated bust target, since the target geometry is not fully convex. We see dark patches in the estimated illumination map where the network cannot correctly estimate the illumination due to self-occlusions. In future work, this limitation may be resolved using a similar method as Verbin et al \cite{verbin2021ref}, in which a learnable ``bottleneck" vector is used to model the target features that are not explained by other parts of the network. 

Secondly, our method is not able to perform re-lighting. While PANDORA can perform diffuse-specular radiance separation, the incident illumination is baked into these radiances and it is chalenging to estimate physically-based material properties, more specifically the material roughness and the diffuse albedo. While our network outputs an $\alpha$ parameter that tunes the rough appearance and models the effect of increasing roughness, such as blurred illumination map (Fig. \ref{fig:suppl_roughness_test}), it does not truly estimate the physics-based roughness. 

%
%

\end{document}